\documentclass{article} 
\usepackage[final]{colm2026_conference}

\usepackage{microtype}
\usepackage{hyperref}
\usepackage{url}
\usepackage{booktabs}
\usepackage{graphicx}
\usepackage{wrapfig}

\usepackage{subcaption}
\usepackage{arydshln}
\usepackage{multirow}
\usepackage{multicol}
\usepackage[most]{tcolorbox}
\usepackage{mdframed}
\usepackage{enumitem}
\usepackage{markdown}
\usepackage[dvipsnames]{xcolor}
\definecolor{musicblue}{HTML}{003990}
\definecolor{soundteal}{HTML}{3AB7A9}
\definecolor{speechpurple}{HTML}{983BB9}
\usepackage{cuted}
\usepackage{capt-of}
\usepackage[most]{tcolorbox}
\usepackage{wrapfig}

\usepackage{lineno}
\usepackage{afterpage}

\definecolor{darkblue}{rgb}{0, 0, 0.5}
\hypersetup{colorlinks=true, citecolor=darkblue, linkcolor=darkblue, urlcolor=darkblue}

\newif\ifshowrev
\showrevfalse

\definecolor{revblue}{HTML}{0B49C4}
\definecolor{revred}{HTML}{C0392B}

\newcommand{\revtag}[1]{%
  \ifshowrev\,{\footnotesize\textcolor{revred}{[#1]}}\fi}
\newcommand{\revtext}[1]{\ifshowrev\textcolor{revblue}{#1}\else#1\fi}
\newcommand{\rev}[2]{\revtext{#2}\revtag{#1}}

\title{\includegraphics[height=1.5em]{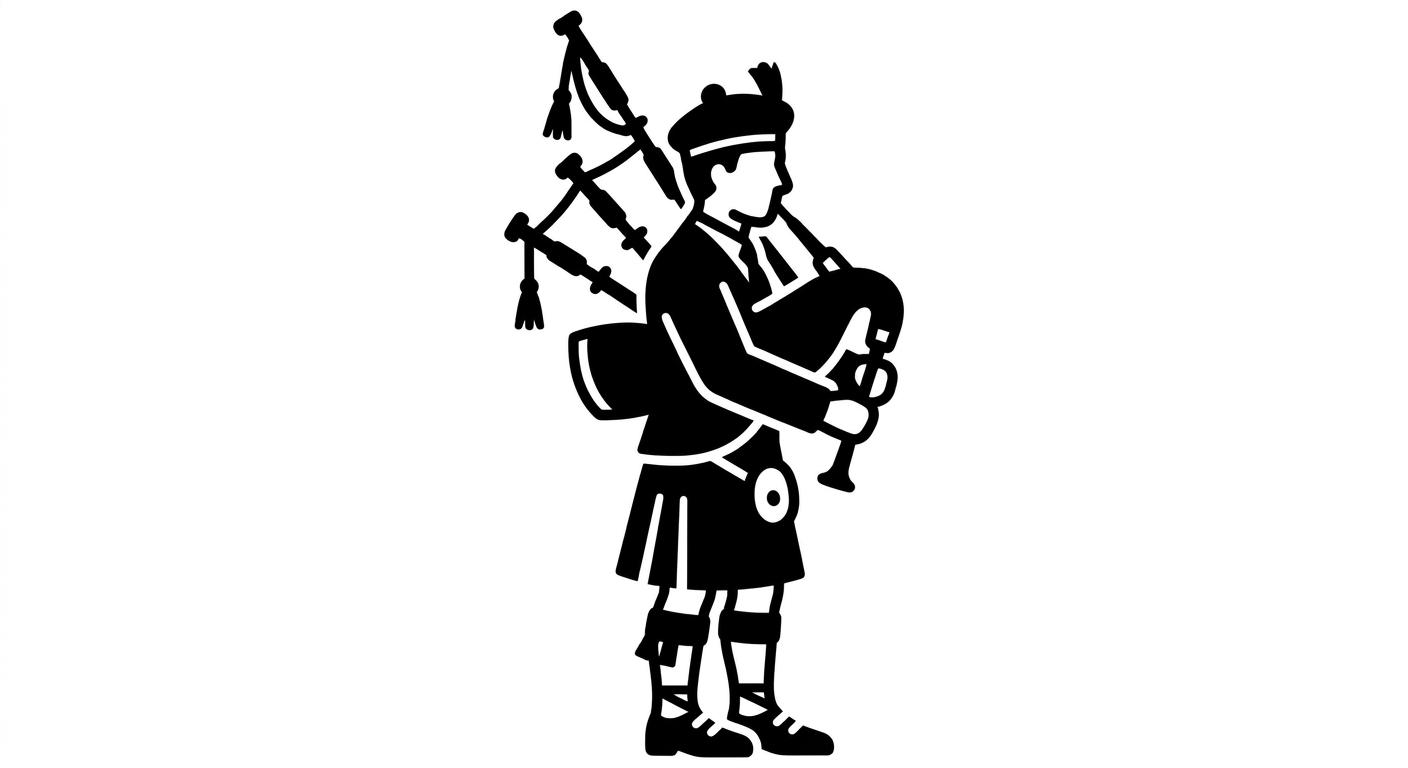} Bagpiper: Solving Open-Ended Audio Tasks \\ \textcolor{white}{spacee} via Rich Captions}

\author{Jinchuan Tian\thanks{Equal contribution.}$^{\ \ 1}$, Haoran Wang\footnotemark[1]$^{\ \ 1}$, Bo-Hao Su\footnotemark[1]$^{\ \ 1}$, Chien-yu Huang\footnotemark[1]$^{\ \ 1}$, Qingzheng Wang\footnotemark[1]$^{\ \ 1}$, \\
\textbf{Jiatong Shi$^{1}$, William Chen$^{1}$, Xun Gong$^{1}$, Siddhant Arora$^{1}$, Chin-Jou Li$^{1}$,} \\
\textbf{Masao Someki$^{1}$, Takashi Maekaku$^{2}$, Keita Goto$^{2}$, Yusuke Shinohara$^{2}$,} \\
\textbf{Jin Sakuma$^{2}$, Chao-Han Huck Yang$^{3}$, Shinji Watanabe$^{1}$} \\
\vspace{2pt} \\
$^{1}$Language Technologies Institute, Carnegie Mellon University \\
$^{2}$LY Corporation \quad $^{3}$NVIDIA \\
\texttt{jinchuat@andrew.cmu.edu}
}

\begin{document}

\ifcolmsubmission
\linenumbers
\fi

\maketitle

\begin{abstract}
Current audio foundation models typically rely on rigid, task-specific supervision (\textit{e.g.}, speech recognition), addressing isolated factors of audio rather than the whole. In contrast, human processes audio holistically, seamlessly bridging raw audio waveform with abstract cognitive concepts (\textit{e.g.}, all perception details of audio events) to execute complex tasks. Grounded in this philosophy, we introduce \texttt{Bagpiper}, an 8B audio foundation model that interprets physical audio via \textit{rich captions}, \textit{i.e.}, comprehensive natural language descriptions that encapsulate the critical cognitive concepts inherent in the audio. By pre-training on a massive corpus of 600B tokens, the model establishes a robust bidirectional mapping between raw audio and this high-level conceptual space. During fine-tuning, \texttt{Bagpiper} adopts a \textit{caption-then-process} workflow, simulating an intermediate cognitive reasoning step to solve diverse tasks without knowing prior task-specific practice. 
Experimentally, \texttt{Bagpiper} achieves universal generation that can uniformly generate speech, sound effects, music, and their arbitrary combinations. It also maintains comparable performance with the 7B Qwen-2.5-Omni for audio understanding.
To the best of our knowledge, \rev{5Apc, R2}{\texttt{Bagpiper} is among the first works that achieve open-ended audio understanding and generation on speech, sound, and music.}
Model, data, and code will be released at \href{https://bagpiper-cmu.github.io/}{\texttt{Bagpiper} Home Page}.
\end{abstract}

\section{Introduction}
\begin{figure}[h]
    \centering
    \includegraphics[width=\linewidth]{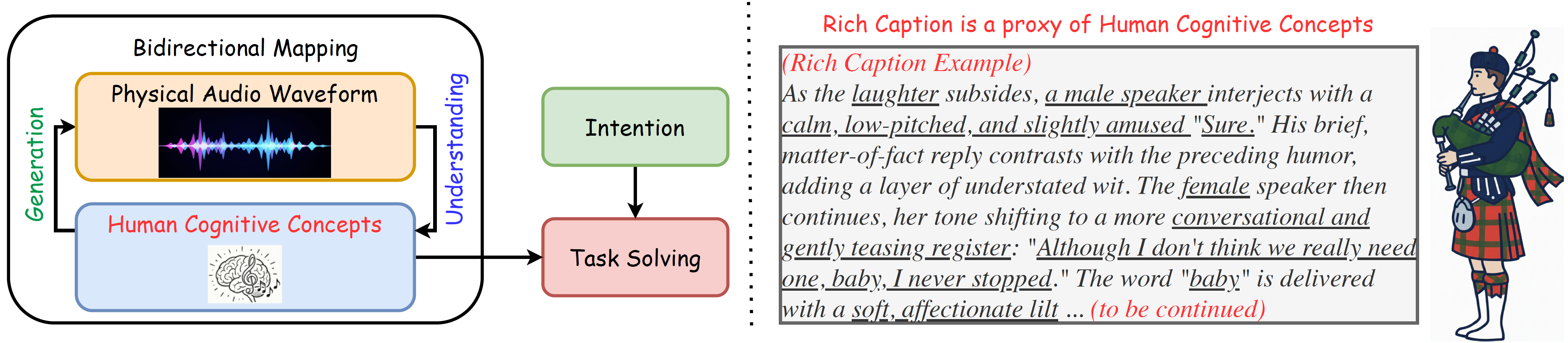}
    \caption{(Left) When solving an audio task, \texttt{Bagpiper} always materializes \textcolor{red}{human cognitive concepts} for the real physical audio waveforms. 
    It produces a subsequent text answer (\textcolor{blue}{understanding}) or audio clip (\textcolor{ForestGreen}{generation}) based on those cognitive concepts.
    (Right) In this work, we take \textcolor{red}{rich caption} as the proxy of human cognitive concepts.}
    \label{fig:concept}
\end{figure}

Current audio foundation models are suboptimal for solving open-ended user requests.
These models mainly rely on massive multi-tasking \citep{af3, yang2023uniaudio, wang2024speechx, abouelenin2025phi, valle2025fugatto, radford2023robust, peng2024owsm, maiti2024voxtlm}, where each task only addresses an isolated aspect of the audio. Additionally, the audio modeling (especially the generation modeling) \citep{du2025cosyvoice, hung2024tangoflux, mousavi2025discrete, shi2024espnet} often has a clear split of speech, music, and sound effects. The lack of a holistic concept at the task- and category-level creates a fundamental bottleneck for scalability: addressing the wide spectrum of audio tasks via exhaustive, domain-specific engineering is unsustainable. More importantly, it generalizes poorly to the user requests that are naturally compositional, flexible, and are thus not well defined.
Thus, we argue for a paradigm shift toward a universal modeling philosophy that unifies all audio tasks and types within a single modeling paradigm.

Human intelligence perceives (understands) and produces~(generates) audio in a holistic, open-ended, and multi-perspective manner~\citep{galantucci2006motor}, seamlessly bridging distinct categories like speech, music, and sound effects \citep{intro1, intro2}.
We propose that audio in this context is fundamentally twofold, defined by both the physical audio signals and the cognitive concepts they evoke in the human mind, such as text transcription, emotion, prosody, sound events, and music genres.
Human intelligence processes audio by bridging these two planes: it converts audio signals into cognitive concepts to understand the world and, conversely, formulates responses derived from these concepts~\citep{o2014auditory}. In this work, we identify the \textit{rich caption} \citep{ma2025omni, anonymous2025videosalmonn, fiscus2006rich, lu2026desta2} as an ideal proxy for this wide range of human cognitive concepts. As illustrated in Figure~\ref{fig:concept}, rich captions provide lengthy, comprehensive natural language descriptions that cover the critical cognitive concepts inherent in an audio signal, offering sufficient information to support diverse downstream open-ended task solving. Guided by this philosophy, the core objective of our audio foundation model, \texttt{Bagpiper}, is to uniformly solve audio tasks by establishing and leveraging a broad, holistic, and robust bidirectional mapping between physical audio signals and these cognitive concepts represented by rich captions. 
In implementation, we build this mapping during pre-training using a massive, curated dataset of 600B tokens, comprising audio-rich caption pairs that span speech, music, and sound effects. We then proceed to a fine-tuning stage where we simulate diverse open-ended request-response pairs with thinking traces, teaching the model how to infer appropriate answers from predicted rich captions~(understanding) and how to translate user requests into rich captions as blueprints before physically producing audio~(generation). Throughout this development pipeline, we incorporate no prior knowledge of specific audio tasks or types; nonetheless, the model demonstrates the capability to solve a diverse array of open-ended audio tasks.

Experimentally, we first validate the efficacy of our pre-training \texttt{Bagpiper}-Base, demonstrating that physical audio and rich captions can be translated bidirectionally with high fidelity across a suite of probing tasks.  
Regarding generative tasks, \texttt{Bagpiper} significantly outperforms prior state-of-the-art text-to-audio models, TangoFlux \citep{hung2024tangoflux} and AudioLDM2-Large \citep{liu2024audioldm}, in an extensive A/B testing on complex generation prompts. Beyond these quantitative metrics, we observe emergent model behaviors that transcend the scope of current evaluation protocols; we invite readers to explore these qualitative examples on \href{https://bagpiper-cmu.github.io/}{\texttt{Bagpiper} Home Page}. 
In terms of open-ended understanding, our \texttt{Bagpiper} achieves comparable capabilities with its 7B competitor Qwen-2.5-Omni~\citep{xu2025qwen2} and AudioFlamingo3~\citep{af3}, \revtext{with competitive results on both AIR-Bench \citep{yang2024air} and AudioBench \citep{wang2025audiobench}.}
To facilitate reproducibility and future research, we commit to releasing our code, data, and trained models.

\section{Bagpiper}\label{sec:method}
\subsection{Architecture} \label{sec:arch} 

\begin{wrapfigure}{r}{0.45\textwidth}
    \centering
    \includegraphics[width=0.45\textwidth]{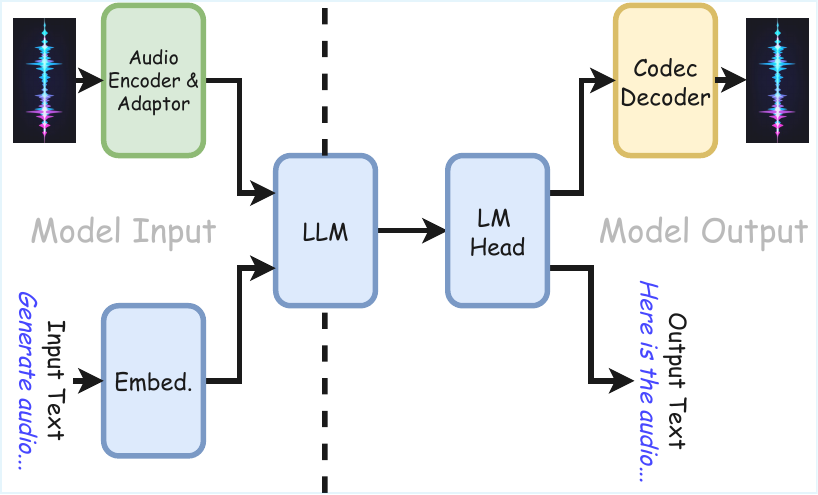}
    \caption{\texttt{Bagpiper} architecture}
    \label{fig:arch}
\end{wrapfigure}

Like \citet{tian2025ualm}, our \texttt{Bagpiper} is designed to process any interleaved audio and text sequences within a unified framework. As shown in Figure \ref{fig:arch}, the backbone is initialized with the decoder-only LLM, i.e., \texttt{Qwen3-8B-Base} \citep{yang2025qwen3}. We adopt the established \textit{Encoder-Adaptor-LLM} architecture \citep{liu2023llava, af3} for audio input, which connects a pre-trained acoustic encoder to the LLM backbone via an MLP layer. 
For audio output, the model auto-regressively predicts multi-stream audio codec tokens, which are then detokenized into audio waveforms.\footnote{The audio encoder and codec model are adopted from \texttt{Qwen3-Omni-30B-A3B} \citep{qwen3omni} and \texttt{X-Codec} \citep{xcodec}, respectively. Each audio is represented by 8 tokens, and the audio token sequences are delay-interleaved \cite{delay}.} 
This design enables the consistent and seamless processing of multimodal inputs and outputs.
For all training stages, we apply standard next-token prediction loss on target tokens only. During inference, we have separate inference configurations for text and audio generation. 
We additionally apply classifier-free guidance \citep{cfg, hussain2025koel} for all audio generation tasks. No noticeable performance change is observed by varying the inference configurations. \rev{9r2o, R1}{Architecture is not our major novelty.} Detailed configurations are in the tables \ref{tab:training_details} and \ref{app:inference_details}.

\subsection{Pre-Training} \label{method_pt}

The central premise of \texttt{Bagpiper} pre-training is that physical audio signals and cognitive concepts (i.e., rich captions) can be mutually translated with high fidelity.
To realize this, our pre-training is to construct a robust bidirectional mapping between audio signals and rich captions via massive-scale learning.
Additionally, it also requires preserving the text processing capabilities of the underlying LLM to support downstream complex task solving.

\subsubsection{Data Curation}
\begin{figure}[t]
    \centering
    \includegraphics[width=\linewidth]{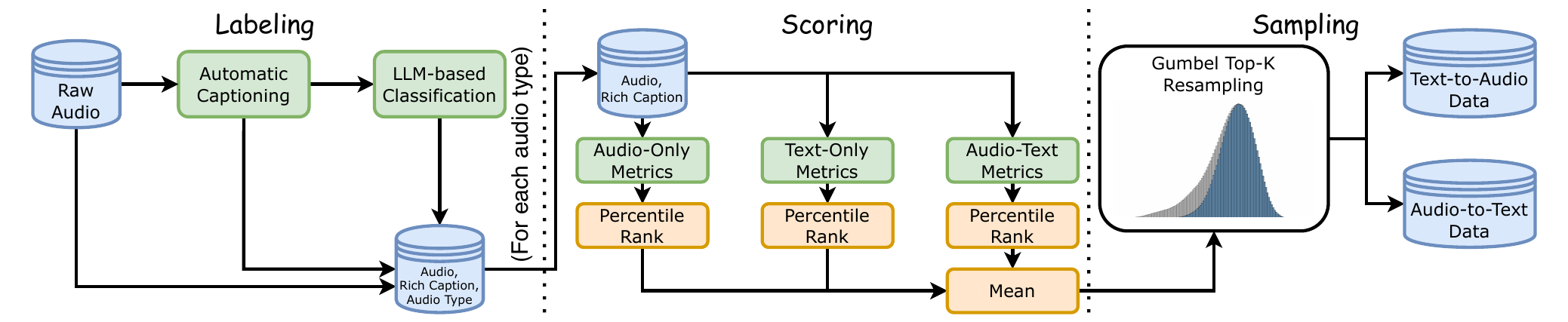}
    \caption{Pre-training data labeling and filtering workflow.}
    \label{fig:filtering}
\end{figure}

\textbf{Collection and Labeling.} To ensure universal audio processing capabilities, we aggregate a massive collection of publicly available audio data covering speech, music, and sound effects (full list in Appendix \ref{app:data_list}). Notably, this mixture features diverse in-the-wild distributions, naturally encompassing complex compositions such as speech with background noise or musical accompaniment. We segment all audio into clips of up to 30 seconds. As in Fig.\ref{fig:filtering}, we generate a rich caption for every clip using the \texttt{Qwen3-Omni-30B-A3B-Captioner} \citep{qwen3omni}. This process yields a total of 422M raw audio-caption pairs.

\textbf{Audio Taxonomy Classification.} As distinct audio categories (speech, music, and sound effects) exhibit divergent quality distributions and severe data imbalances, a stratified processing approach is used to keep a rough balance in the curated pre-training mixture.\footnote{
  E.g., we observe the average aesthetic score of \textit{music} is much higher than \textit{sound effects} on average.
}
\rev{5Apc, R1}{While audio classification directly from waveforms has been studied extensively~\citep{gemmeke2017audioset, kong2020panns, gong2021ast, chen2023beats}, categorizing audio via rich captions remains underexplored, and we find it empirically effective.} We explicitly categorize all audio clips into speech, music, or sound effects by prompting \texttt{Qwen3-32B} to analyze their rich descriptions. This text-based strategy proves highly effective, achieving a 93\% accuracy on the MMAU dataset \citep{sakshi2024mmau} when benchmarked against the original ground-truth category labels.

\textbf{Stratified Data Filtering and Sampling.} \label{method_data_curation}
For each audio category, we score the audio-text pairs along three dimensions: (1) Audio-Only: We use UTMOS \citep{saeki2022utmos, shi2025versa} for speech and audiobox-aesthetics \citep{aes_score} for music and sound effects. (2) Text-Only: We apply heuristic filters combined with an LLM-as-a-judge approach \citep{zheng2023judging}, evaluating three specific rubrics \citep{peng2025dataman}. (3) Audio-text alignment: We compute the semantic similarity between the audio and caption using a CLAP-based model\footnote{Due to the context length limit of the CLAP model, we summarize them into short captions.} \citep{laionclap2023}.
To unify these diverse metrics, we compute the percentile rank of each sample for every dimension within its specific category and average these percentiles to form a final quality score. Next, to maintain data diversity while removing low-quality outliers, we employ a Gumbel Top-$k$ resampling trick \citep{kool2019stochastic} to perform a soft truncation of the tail distribution. Finally, we perform the text-based de-duplication by MinHash \citep{666900} using a setup as in \citet{li2024datacomplm}.
We observed that the generated rich captions possessed high natural diversity, resulting in the removal of only $\sim$1M pairs during de-duplication.
Further details are provided in Appendix \ref{app:curation}.

\textbf{Text-Only Corpus.} To preserve the text capability of the base LLM, we incorporate high-quality text-only corpora into the pre-training mixture. We specifically utilize datasets from \cite{bercovich2025llama} and the mid-training data of \cite{olmo2025olmo}, totaling 150B tokens. These datasets are yielded from harsh curation in prior works, so we skip further curation on text-only data.

\textbf{Training Budget and Scheduling.} Like in \citep{maiti2024voxtlm, olmo2025olmo}, we first allocate a total training budget of 600B tokens.\footnote{A token represents either a text subword or an audio frame.}
Following empirical observations from \citet{tian2025ualm}, we adopt a data ratio of 300B (Text-to-Audio) : 150B (Audio-to-Text) : 150B (Text-only). This distribution serves two purposes: first, \citet{tian2025ualm} finds that audio generation (Text-to-Audio) typically requires a longer training horizon to converge than understanding; second, it perfectly balances the output modalities, ensuring a 1:1 ratio between text-output and audio-output supervision. Within the text-to-audio and audio-to-text tasks, we evenly split the budget across speech, music, and sound effects, which effectively up-samples the music and sound effects data by a factor of 3--4$\times$ to counteract the natural scarcity of these domains compared to speech.

\subsubsection{Training Strategy} 
We implement a two-stage training protocol following \cite{wu2025janus}. Stage 1 begins with an adapter and embedding warm-up stage (2k steps), where the audio encoder and LLM backbone are frozen; only the embeddings and the adaptor are optimized. During this stage, we employ a substantially larger batch size and learning rate compared to the subsequent stage to rapidly align the modalities. Following this, stage 2 launches the full pre-training stage. Here, we unfreeze the relevant parameters and apply a linear learning rate warm-up followed by cosine decay. 
The loss of audio tokens is observed to be much higher than that of text tokens on average.
Consequently, we use sequence packing \citep{krell2021efficient} that carefully fuses samples from both sides to reduce the training loss fluctuation.
Detailed training hyperparameters are provided in table \ref{tab:training_details}. We refer to the resulting model as \texttt{Bagpiper}-Base.

\subsection{Supervised Fine-Tuning (SFT)} \label{method_sft}
In this SFT stage, we evolve \texttt{Bagpiper}-Base into a generalist problem solver. Consistent with our goal of overcoming the limitations of isolated supervision, we deliberately eschew traditional categorical labels during data simulation. Instead, we condition Bagpiper to treat rich captions as a universal semantic interface. By grounding the model’s reasoning process entirely in these natural language descriptions, we enable it to handle the boundless variety of open-ended tasks described in our introduction.

\begin{figure*}[t]
    \centering
    \includegraphics[width=0.9\textwidth]{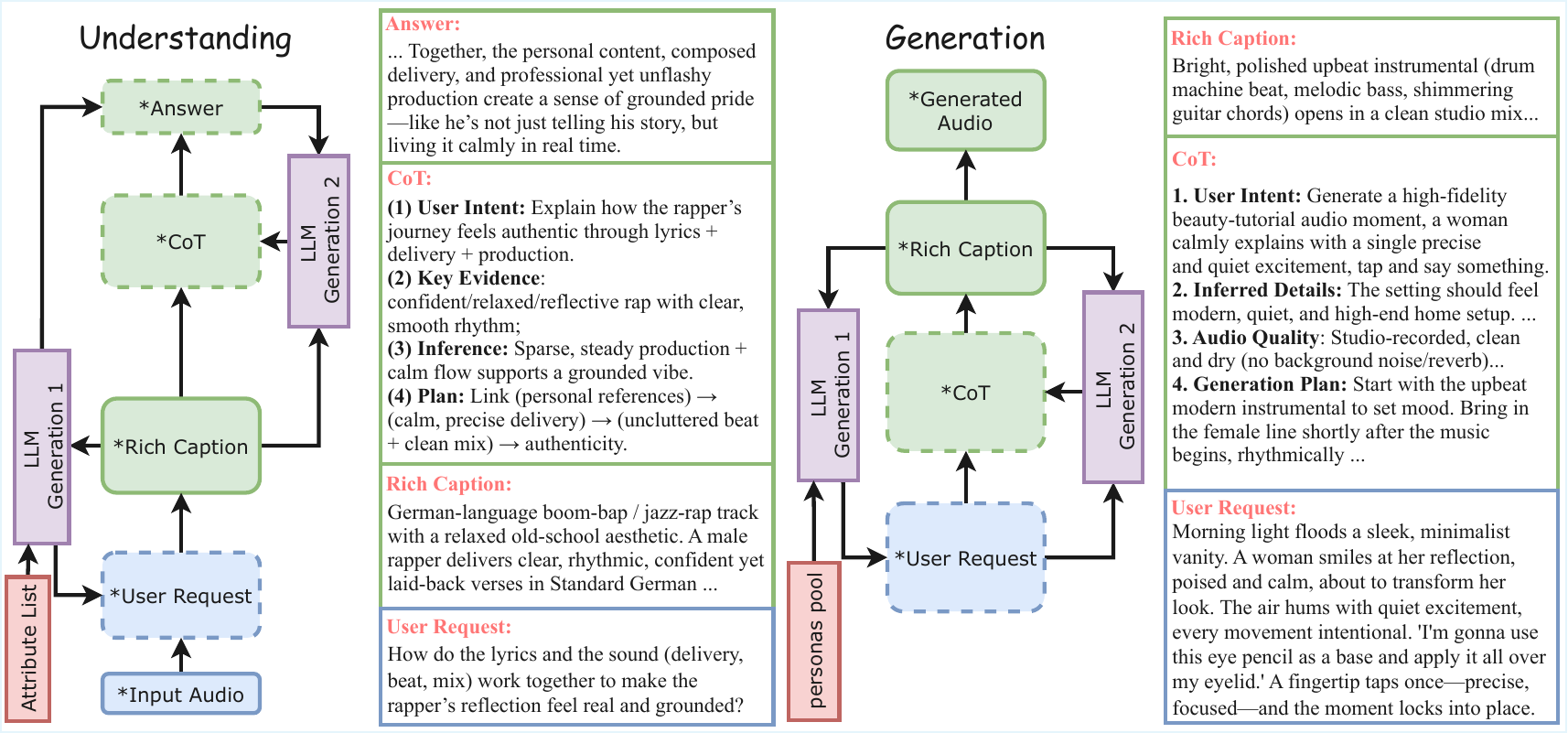}
    \caption{ SFT data simulation pipeline \S\ref{sft_pipeline} for open-ended audio task solving. \textcolor{blue}{Blue} means model input; \textcolor{ForestGreen}{green} means model output. Dashed boxes are simulated by prompting LLMs. Components with * are included in training sequences.}
    \label{fig:sft_pipeline}
\end{figure*}

\subsubsection{Data Collection and Gemini Captioning}
We curate diverse and high-quality audio-caption pairs for both audio understanding and generation.
For audio understanding, we sample $\approx$271k examples from a series of open datasets. We use our manual prompt to query \texttt{Gemini-3-Flash/Gemini-2.5-Pro} to obtain \rev{xDgL, Q3}{the rich captions, which are JSON-structured and more suitable for downstream data simulation} (detailed distributions can be found in Appendix~\ref{app:data_list}).
For audio generation, we continue our strategy of Gumbel Top-$k$ sampling, but with a much lower temperature to select the highest quality data in our pre-training. We end up with 1M samples. 
Our audio generation data relies on the pre-training captions rather than \texttt{Gemini}-generated ones, preventing data leakage as \texttt{Gemini} is later employed as the evaluator in \S\ref{exp_sft_gen}.

\subsubsection{Fine-Tuning Data Construction} \label{sft_pipeline}
\textbf{Pipeline: Caption-then-Process.} 
Our data simulation always starts from audio-caption pairs, and uniformly enforces an explicit ``thinking'' pattern.
For \textbf{audio understanding}, we adopt the template: \textit{[Audio, User Request, Rich Caption, CoT, Answer]}. As shown on the left of Figure \ref{fig:sft_pipeline}, we prompt a strong text LLM to first synthesize a valid \textit{(User Request, Answer)} pair given the ground-truth rich caption, and subsequently generate a \textit{Chain-of-Thought (CoT)} trace \citep{wei2022chain} that reasons from cognitive concepts (i.e., rich captions) to the expected answer.
For \textbf{audio generation}, we adopt the template: \textit{[User Request, CoT, Rich Caption, Audio]}. As shown in the right of Figure \ref{fig:sft_pipeline}, the simulation process is similar: the LLM generates a creative \textit{User Request} based on the rich caption, followed by a \textit{CoT} trace that plans the audio content before outputting the caption.

\textbf{Diversification.} To cover more diversity of user requests, we employ targeted prompting strategies using strong open-source text LLMs.\footnote{Specifically \texttt{Qwen3-235B-A22B-Instruct-2507-FP8} and \texttt{GPT-OSS-120B}.} For \textbf{understanding}, we sample from an attribute list of 108 attributes\footnote{
  We randomly sample 10k pre-training captions, ask \texttt{Gemini-3-Pro} to analyze them, and obtain this attribute list \rev{JEWz, Q2}{(App.~\ref{app:taxonomy})}.
} to construct complex, multi-faceted inquiries. The LLMs are instructed to generate complex but reasonable requests with self-selected attributes. Our simulation also includes the cases where multiple related attributes are queried together to form compositional requests \citep{anonymous2025long}. 
For \textbf{generation}, we adopt a ``Persona-based'' prompting strategy, selecting from 12 distinct user personas \citep{ge2024scaling} to generate varying generation requests.
Besides the user requests that explicitly mention the exact audio events and details, we also generate the imaginary user requests that describe a feeling, an atmosphere, or a scene, which potentially boost the model's creativity in audio generation \citep{tian2025ualm}.

\textbf{Automatic Quality Assurance.}
To ensure high data quality, we implement a rigorous filtering pipeline utilizing the LLM-as-judge framework \citep{zheng2023judging} for descriptive quality estimation~\citep{chenaudio}. We employ \texttt{Qwen3-235B-A22B-Instruct-2507-FP8} to evaluate each sample on a 5-point scale, assessing five key dimensions: user request diversity, request-response alignment, thinking trace coherence, rich caption quality, and overall training value. By retaining only those samples with an average score exceeding 3, we distill our simulated data into a final set of $845$k understanding samples and 1.47M generation samples. \rev{5Apc, Q4}{We validate this cutoff in App.~\ref{app:filter_validation}.} Note that multiple requests will be generated for each audio-caption pair before this filtering.\footnote{\revtext{Within each of the understanding and generation tracks, the evaluation judge participated in neither caption generation nor SFT data simulation. See App.~\ref{app:model_roles} for details.}}\revtag{5Apc, R4}

\section{Experiments}
\subsection{Pre-Training Evaluation Setup} \label{exp_pt}
The \texttt{Bagpiper}-Base is not for real task-solving. Instead, our evaluation design for this pre-trained model is to quantify how much critical cognitive concepts are preserved in the rich captions that can support downstream task solving. We mainly rely on LibriSpeech Test-Clean \citep{librispeech} for local, phonetic-level capability and MMAU-Mini \citep{sakshi2024mmau} for other audio-related capability in general, which covers speech, music, and sound effects.

\textbf{Audio Understanding (Information Retention).} 

Following \citet{ma2025omni}, to quantify how much cognitive concepts are preserved for downstream understanding, we feed the generated caption into \texttt{Gemini-3-Flash} and request it to perform audio tasks solely based on the text, without access to the original audio. We report the Word Error Rate (WER) on LibriSpeech Test-Clean and question-answering accuracy on MMAU-Mini. High performance here indicates that the captions generated from audio successfully encapsulate the essential semantics of the audio clips for understanding. 

\textbf{Audio Generation (Reconstruction Fidelity).} 
We evaluate whether the pre-trained model can physically produce the audio waveform with the audio caption generated by the original captioner \citep{qwen3omni}.
We generate audio conditioned solely on the rich captions and compare the output to the ground truth audio. We report WER on LibriSpeech Test-Clean to assess phonetic preservation.\footnote{
 We only test the examples whose rich captions contain the fully correct text transcriptions, which is around 60\% of the volume of the whole test set.
}
For general audio, we employ a model-based metric on MMAU-Mini: We use \citet{shi2025versa} to compute the FAD \citep{kilgour2018fr} and audio-audio CLAP similarity \cite{laionclap2023} between the reference and generated audio, and average the number across all three categories.

\textbf{Cycle Consistency.} 
Given the unified understanding-generation nature of our model, isolated evaluation is often insufficient. We therefore follow \citet{hori2019cycle} to conduct a cycle consistency test to verify the robustness of the bidirectional mapping, which is similar to evaluating an auto-encoder. We perform consecutive \textit{Audio $\to$ Rich Caption $\to$ Audio} (reconstruction) and \textit{Rich Caption $\to$ Audio $\to$ Rich Caption} (grounding) loops using \texttt{Bagpiper}-Base. We measure the cosine similarity between the input and the final output using \texttt{Qwen3-Embedding-8B} \citep{yang2025qwen3} for the text domain (\textit{Text Sim.}) and the CLAP audio encoder \citep{laionclap2023} for the audio domain (\textit{Audio Sim.}). High similarity confirms minimal information loss through mutual translation between the physical audio signals and cognitive concepts.

\textbf{Text Capability Preservation.} To verify that incorporating audio objectives does not degrade language reasoning, we evaluate \texttt{Bagpiper}-Base on standard text benchmarks: MMLU-Redux (world knowledge) \citep{mmlu_redux}, GPQA-Diamond (general reasoning) \citep{rein2024gpqa}, and GSM8K (mathematical reasoning) \citep{cobbe2021training}.

\subsection{Pre-Training Results and Analysis}

\begin{wraptable}{r}{8cm}
    \centering
    \vspace{-10pt}
    \caption{Understanding probing results for \texttt{Bagpiper}-Base. Both models are NOT for real task solving; the task is solved by an external LLM using the generated rich captions.}
    \scalebox{0.7}{
    \begin{tabular}{lccc}
    \toprule
        Model & Param. & WER ($\downarrow$) & MMAU-Mini ($\uparrow$) \\
        \midrule
        Qwen3-Captioner & 30B-A3B  & 5.5 & \textbf{71.1}   \\
        Bagpiper-Base (ours) & 8B   & \textbf{5.0} & 69.0   \\
        \bottomrule
    \end{tabular}}
    \label{tab:pt_und}
\end{wraptable}
In \textbf{audio understanding}, our objective is to match the performance of our teacher model. As shown in Table \ref{tab:pt_und}, \texttt{Bagpiper}-Base (8B) achieves performance parity with the \texttt{Qwen3-Captioner} (30B) topline \citep{qwen3omni}. \rev{xDgL, R2; JEWz, R3}{This comparison is a distillation-parity sanity check rather than evidence that \texttt{Bagpiper}-Base exceeds its caption teacher.}
This result validates the effectiveness of our distillation efforts from the captioner, demonstrating that our 8B backbone successfully learns to extract comprehensive cognitive details comparable to much larger models.
Also, we find that both models show considerable hallucination on the ASR task (5.5\% and 5.0\%). After manually checking, the hallucination is almost all from the captioner models, not the probing LLM. This is further discussed in table \ref{tab:sft_und}. Overall, these probing results conclude that physical audio signals can be effectively compressed into cognitive concepts by \texttt{Bagpiper}-Base.

\begin{wraptable}{r}{8cm}
    \centering
    \vspace{-10pt}
    \caption{Generation probing results. Our model is prompted by rich captions, while others are prompted by text transcriptions or short captions.}
    \scalebox{0.7}{
    \begin{tabular}{lcccc}
    \toprule 
      && Text-to-Speech & \multicolumn{2}{c}{Text-to-Audio} \\ 
        Model & Param. & WER ($\downarrow$) & FAD ($\downarrow$) & CLAP ($\uparrow$) \\
        \midrule
        OpusLM (TTS) & 8B &  4.0 & - & - \\
        CosyVoice3 (TTS) & 0.5B & 2.9 & - & - \\
        TangoFlux (TTA) & 0.5B & -  & 5.15 & 0.50\\
        AudioLDM2-L (TTA) & 1.5B & - & 3.79 & 0.48 \\
        \hdashline
        Bagpiper-Base (ours) & 8B  & \textbf{1.8} & \textbf{2.98} & \textbf{0.55} \\
        \bottomrule
    \end{tabular}}
    \label{tab:pt_gen}
\end{wraptable}
For \textbf{audio generation}, to the best of our knowledge, there are no rich caption-to-audio baselines to compare with. Thus, we compare our \texttt{Bagpiper}-Base with specialized state-of-the-art baselines in speech (TTS; OpusLM~\citep{tian2025opuslm} and CosyVoice3~\citep{du2025cosyvoice}) and sound/music (TTA; TangoFlux~\citep{hung2024tangoflux} and AudioLDM2-Large~\citep{liu2024audioldm}), respectively. Note that these baselines are limited to their standard input format (transcripts for TTS; short captions for TTA), 
whereas \texttt{Bagpiper}-Base utilizes the full rich captions. 
Table \ref{tab:pt_gen} shows that \texttt{Bagpiper}-Base achieves WER comparable to dedicated TTS systems and outperforms TTA baselines in fidelity. This suggests that cognitive concepts in rich captions can be accurately grounded back into physical audio signals with \texttt{Bagpiper}-Base.

\begin{wraptable}{r}{8cm}
    \centering
    \vspace{-10pt}
    \caption{Cycle consistency probing results}
    \scalebox{0.7}{
    \begin{tabular}{cccc}
    \toprule
     Audio-to-Text & Text-to-Audio &  Audio-Sim. ($\uparrow$)  & Text-Sim. ($\uparrow$) \\
     \midrule
     Qwen3-Omni & AudioLDM2-L & 0.445 & 0.465\\
     Qwen3-Omni & TangoFlux & 0.361 & 0.486\\
     AudioFlamingo3 & AudioLDM2-L & 0.457 & 0.515\\
     AudioFlamingo3 & TangoFlux & 0.471 & 0.570\\
     \hdashline
     \multicolumn{2}{c}{Bagpiper-Base (ours)} & \textbf{0.502} & \textbf{0.840} \\
     \bottomrule
    \end{tabular}}
    \label{tab:pt_cycle}
\end{wraptable}
Subsequently, on the \textbf{cycle consistency} test (Table \ref{tab:pt_cycle}), we compare \texttt{Bagpiper}-Base against pipelined combinations of different understanding \citep{qwen3omni, af3} and generation models \citep{hung2024tangoflux, liu2024audioldm}. Since baseline generation models cannot process the lengthy rich captions, \rev{JEWz, Q3}{we summarize captions only for these baselines to fit their context windows; \texttt{Bagpiper}-Base receives the full captions.} \texttt{Bagpiper}-Base significantly outperforms these composite pipelines in both audio and text similarity, confirming that a unified model establishes a far more coherent bidirectional mapping than loosely coupled systems.

\begin{wraptable}{r}{8cm}
    \centering
    \vspace{-10pt}
    \caption{Text capability probing results}
    \scalebox{0.63}{
    \begin{tabular}{lccc}
    \toprule
         & MMLU-Redux ($\uparrow$) & GPQA-Diamond ($\uparrow$) & GSM8K ($\uparrow$) \\
    \midrule
        Qwen3-8B-Base & \textbf{76.1} & 39.3 & \textbf{89.8} \\
        Bagpiper-Base (ours) & 74.3 & \textbf{41.9} & 88.2 \\ 
    \bottomrule
    \end{tabular}}
    \label{tab:pt_text}
\end{wraptable}
As demonstrated in Table \ref{tab:pt_text}, \texttt{Bagpiper}-Base maintains close performance to its initialization checkpoint (\texttt{Qwen3-8B-Base}) across knowledge, general reasoning, and math benchmarks. Preserving this strong textual foundation is not merely for maintenance; it directly powers our open-ended task solving during downstream fine-tuning. By retaining the model's inherent logic and world knowledge, \texttt{Bagpiper} can leverage complex text-based reasoning (e.g., Chain-of-Thought) to decompose and solve flexible audio instructions within the text domain and reflect it back to the audio modality.

\subsection{SFT Evaluation}
Unlike prior probing tests in \S\ref{exp_pt}, this SFT evaluation is to validate the fine-tuned \texttt{Bagpiper}'s efficacy in addressing real audio tasks through a unified, open-ended interface.

\subsubsection{Audio Understanding} \label{exp_sft_und} 
\begin{table}[t]
    \centering
    \caption{Audio understanding results after fine-tuning. $^\dagger$Qwen3-Omni-Instruct is a non-size-matched reference; $^\ddagger$UALM results are paper-reported rather than re-run.\revtag{xDgL, R3/R4; JEWz, R1--R3; 5Apc, R6; AC, \#2/\#3}}
    \label{tab:sft_und}
    \begingroup
    \setlength{\tabcolsep}{2.4pt}
    \scriptsize
    \revtext{
    \begin{tabular}{lccccccc}
    \toprule
    Model & Size & WER ($\downarrow$) & MMAU-Mini ($\uparrow$) & MMAU ($\uparrow$) & MMAR ($\uparrow$) & AIR-chat ($\uparrow$) & AudioBench ($\uparrow$) \\
    \midrule
    Audio Flamingo 3 & 8B & 1.6 & 73.3 & \textbf{76.7} & 55.2 & \textbf{6.80} & \textbf{70.45} \\
    Qwen2.5-Omni & 7B & 1.6 & 71.5 & 65.5 & 56.7 & 6.30 & 68.03 \\
    Qwen3-Omni-Instruct$^\dagger$ & 30B-A3B & \textbf{1.2} & \textbf{78.0} & 75.2 & \textbf{70.4} & -- & -- \\
    MiMo-Audio & 7B & 3.7 & 73.7 & 66.0 & -- & -- & -- \\
    Voxtral-Mini & 3B & 1.8 & 61.3 & 57.1 & -- & -- & -- \\
    Voxtral-Small & 24B & 1.5 & 68.8 & 62.2 & -- & -- & -- \\
    UALM$^\ddagger$ & 8B & -- & -- & 74.1 & 55.2 & -- & -- \\
    \hdashline
    Bagpiper (ours) & 8B & 2.5 & 74.5 & 73.1 & 57.0 & 6.57 & 70.39 \\
    \bottomrule
    \end{tabular}}
    \endgroup
\end{table}
We compare recent systems \citep{af3, xu2025qwen2, qwen3omni, mimoaudio, voxtral, tian2025ualm} on LibriSpeech test-clean, MMAU-Mini and full MMAU \citep{sakshi2024mmau}, and MMAR \citep{ma2025mmar}; the final two columns report AIR-Bench-chat \citep{yang2024air} and five AudioBench QA subsets \citep{wang2025audiobench}.\footnote{The five subsets are CN-College-Listen, DREAM-TTS, Public-SG-SpeechQA, WavCaps-QA, and AudioCaps-QA. \revtext{Some evaluation clips overlap with the SFT pool despite test-split exclusion; see App.~\ref{app:data_list} for details.}}\revtag{5Apc, R6; AC, \#2} \rev{xDgL, R3/R4; JEWz, R1--R3}{Table~\ref{tab:sft_und} shows that \texttt{Bagpiper} leads MMAU-Mini among 7--8B systems with complete results and scores above Qwen2.5-Omni, Audio Flamingo 3, and UALM on MMAR. On full MMAU it trails Audio Flamingo 3, the larger Qwen3-Omni-Instruct reference, and UALM.} \rev{AC, \#3}{UALM's numbers are paper-reported: with no official checkpoint released, a protocol-matched comparison is left to follow-up.} \revtext{Notably, \texttt{Bagpiper} learns to solve ASR and open-ended problems through the shared caption-then-process objective, without task-specific heads or prior knowledge of the individual task formats.} \rev{xDgL, R2}{WER falls from 5.0 before SFT to 2.5 due to more calibrated supervision signals; App.~\ref{app:error_propagation} scopes caption-error effects on MMAU-Mini speech.} \rev{xDgL, Q2/Q4/Q5}{Experiments in App.~\ref{app:understanding_ablations} demonstrate that our audio pre-training, CoT-format supervision, and end-to-end caption-then-process design benefit overall system performance; they also show that \texttt{Bagpiper} can be strengthened by scaling up to a 30B-A3B backbone.}

\subsubsection{Audio Generation} \label{exp_sft_gen}
\begin{wraptable}{r}{5cm}
    \centering
    \vspace{-25pt}
    \caption{Text-to-speech results after fine-tuning.\revtag{JEWz, R1}}
    \scalebox{0.8}{
    \begin{tabular}{lc}
    \toprule
         Model &  WER \\
         \midrule
        \revtext{AnyGPT} & \revtext{24.9} \\
        \revtext{SpeechGPT} & \revtext{42.8} \\
        CosyVoice3 & 2.9 \\
        Bagpiper (ours) & \textbf{2.7}\\
        \bottomrule
    \end{tabular}}
    \label{tab:sft_tts}
\end{wraptable}
We examine our model on the classical text-to-speech (TTS) task on Librispeech Test-Clean.
As our model is pre-trained for a general purpose, we prompt the model to generate with \textit{a calm, neutral voice}, plus the text transcription. As in the table \ref{tab:sft_tts}, our generalist \texttt{Bagpiper} outperforms the dedicated TTS model on the WER, even though it has never been tuned for this task. \rev{xDgL, R2}{App.~\ref{app:error_propagation} shows that the intermediate rich caption preserves the requested transcription almost exactly ($0.2\%$ WER), locating the residual errors in the caption-to-audio stage.} \rev{JEWz, R1}{We also list AnyGPT \citep{anygpt} and SpeechGPT \citep{speechgpt}, which pioneered unified speech and audio generation with language models; their intelligibility on this specialized metric is well below the dedicated systems', so the comparison contextualizes current unified-generation TTS quality rather than diminishing these early systems.} \rev{9r2o, Q1/R2}{To compare against specialized TTS systems on their own terrain, App.~\ref{app:tts_specialized} reports a TTS-specialized \texttt{Bagpiper} variant A/B-tested against \texttt{Qwen3-TTS}, \texttt{CosyVoice3}, and \texttt{VibeVoice} under both judges and both rubrics.}

\begin{wrapfigure}{r}{0.48\textwidth}
    \centering
    \vspace{-12pt}
    \includegraphics[width=\linewidth]{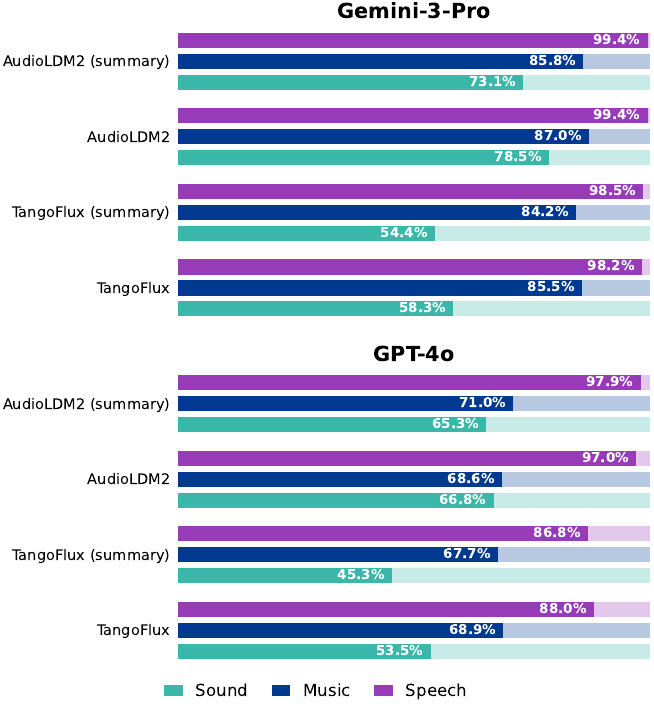}
    \caption{A/B win rates of \texttt{Bagpiper} against text-to-audio baselines under two judges, \texttt{Gemini-3-Pro} and \texttt{GPT-4o}, broken down by \textbf{\textcolor{soundteal}{Sound}}, \textbf{\textcolor{musicblue}{Music}}, and \textbf{\textcolor{speechpurple}{Speech}}. The darker region is \texttt{Bagpiper}'s win rate and the lighter region its loss.\revtag{5Apc, Q3; 9r2o, R2; AC, \#2}}
    \label{fig:win_rate_results}
    \vspace{-10pt}
\end{wrapfigure}
We subsequently evaluate the model's capability for general audio generation, specifically its ability to synthesize arbitrary combinations of diverse audio types. To assess performance on complex instructions, we query \rev{5Apc, Q3; 9r2o, R2}{\texttt{Claude-Opus-4.8} (not used in model development)} to generate detailed prompts for each audio sample in MMAU-Mini, yielding a total of 1,000 evaluation samples.
We compare \texttt{Bagpiper} against state-of-the-art baselines, including TangoFlux \citep{hung2024tangoflux} and AudioLDM2-Large \citep{liu2024audioldm}. To ensure a fair comparison (standard text-to-audio models may struggle with lengthy instructions), 
we also evaluate the baselines using condensed 15-word summaries of the original prompts (denoted as \textit{summary}). \rev{9r2o, Q3}{As a matched-input control, App.~\ref{app:matched_input} additionally gives \texttt{Bagpiper} the same 15-word summary and confirms it still wins in aggregate under both judges.}
We conduct a side-by-side preference evaluation using \texttt{Gemini-3-Pro} \rev{5Apc, Q3; 9r2o, R2; AC, \#2}{and \texttt{GPT-4o}} as judges\rev{5Apc, R4; AC, \#2}{, with the two systems presented in randomized order to control for position bias}.
As shown in Fig.~\ref{fig:win_rate_results}, \rev{5Apc, Q3; 9r2o, R2; AC, \#2}{\texttt{Bagpiper} wins in aggregate in every comparison under both judges.}
The advantage is most pronounced on the \textit{speech} subset, where \texttt{Bagpiper} synthesizes intelligible speech integrated with complex sound events or music, a capability largely absent in prior text-to-audio models. \rev{5Apc, Q3; 9r2o, R2}{Overall, these results indicate that \texttt{Bagpiper} can faithfully follow natural-language user prompts and generate complex, multi-faceted audio.}

\subsection{Qualitative Observation}\label{subsec:qualitative}
As an open-ended task solver, \texttt{Bagpiper} exhibits several attractive behaviors that extend beyond the scope of current evaluation protocols, particularly in the domain of audio generation. We highlight key observations below and refer readers to our \href{https://bagpiper-cmu.github.io/}{\texttt{Bagpiper} Project Page} for comprehensive demonstrations.

\afterpage{%
\begin{figure}[t]
    \centering
    \includegraphics[width=\linewidth]{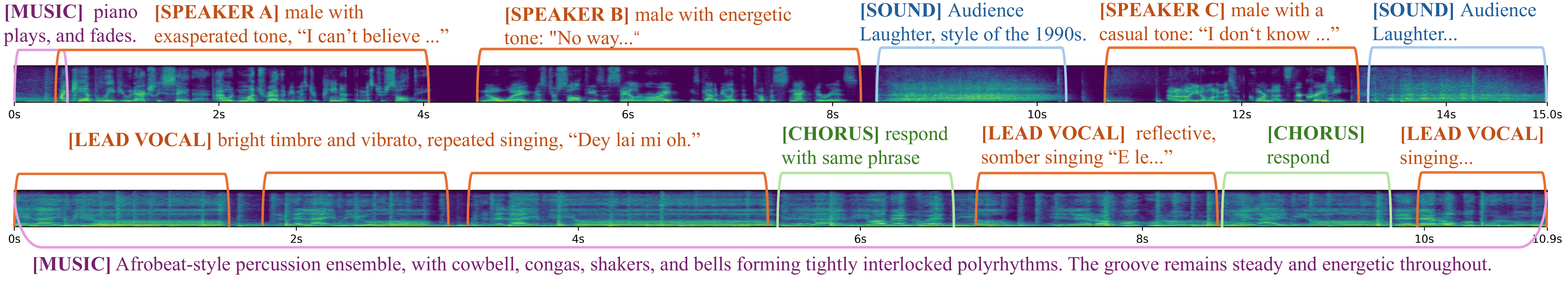}
    \caption{Spectrum of audio samples generated by \texttt{Bagpiper}. Our model is capable of generating multi-speaker dialogue with sound and music (upper) and singing voice with accompaniment (lower).}
    \label{fig:caption_example}
\end{figure}%
}

\textbf{Compositional Audio Synthesis.}
As demonstrated in Fig.\ref{fig:caption_example}, our model is capable of generating complex audio sequences that simultaneously incorporate speech, environment sounds, and music, within a single clip. Notably, this includes conversational features such as multi-speaker dialogue with sound effects.

\textbf{Instruction Following.}
Leveraging its reasoning capabilities, the model processes instructions that extend beyond simple acoustic descriptions to include logic-based constraints. For instance, when prompted to \textit{"sing 'Merry Christmas' twice"}, the model correctly produces the repetition: \textit{"Merry Christmas, Merry Christmas"}.

\textbf{World Knowledge Utilization.}
The model effectively leverages world knowledge derived from its textual pre-training to ground audio generation in specific contexts. For example, when prompted to \textit{"Imagine walking through Disneyland"}, the model generates contextually appropriate audio, such as marching bands and fireworks, without needing explicit acoustic descriptions of those elements\rev{JEWz, R5}{, with further discussion of these qualitative observations in App.~\ref{app:stage_study}}.

\section{Related Work}
\rev{5Apc, R2/R3/Q7; JEWz, R1/R2; xDgL, R3}{We outline the closest lines below; App.~\ref{app:related_work} gives the comprehensive discussion.}

\textbf{Open-Ended Task Solving.} Text foundation models attain broad generalization through massive multi-task supervision \citep{flant5}, but replicating this task scaling in audio is impractical, and existing audio language models still struggle with compositional instructions outside their training distribution. \texttt{Bagpiper} instead offloads open-ended reasoning to the text LLM through rich captions, needing no explicit audio-task supervision.

\textbf{Rich Captions as a Modal Interface.} Detailed captions have been shown to enhance individual downstream capabilities, but prior work remains task-specific, small-scale, and unidirectional. \texttt{Bagpiper} instead establishes a universal, bidirectional mapping between natural language and all audio categories.

\textbf{Universal Audio Processing.} While universal audio understanding has advanced, audio generation remains fragmented across models specialized for speech, sound, or music \citep{hung2024tangoflux}. \texttt{Bagpiper} synthesizes integrated acoustic scenes with arbitrary, overlapping speech, music, and sound events in a single clip.

\section{Limitations}
\revtext{\texttt{Bagpiper} depends on an intermediate rich caption, so both understanding and generation inherit the caption model's errors, and residual ASR or captioner hallucinations can propagate downstream (App.~\ref{app:error_propagation}).}\revtag{xDgL, R1/R2} \revtext{Its attribute taxonomy and SFT data are LLM-agent-generated rather than expert-verified,}\revtag{JEWz, Q2} \revtext{and evaluation relies heavily on LLM-as-judge.}\revtag{JEWz, R4; AC}

\section{Conclusion}
We presented \texttt{Bagpiper}, an 8B model that uses rich captions as a shared interface for audio understanding and generation across speech, sound, and music. Pre-training establishes a bidirectional mapping between audio signals and cognitive concepts expressed as rich captions, while caption-then-process SFT supports open-ended tasks. The reported results show competitive understanding and flexible mixed-audio generation, subject to the evaluation and contamination caveats above.

\section*{Acknowledgments}
Parts of this work used the PSC Bridges2 system and Delta/DeltaAI system at NCSA through allocations CIS210014 and IRI120008P from the ACCESS program, supported by NSF grants \#2138259, \#2138286, \#2138307, \#2137603, and \#2138296.

\bibliography{colm2026_conference}
\bibliographystyle{colm2026_conference}

\newpage
\appendix

\newtcolorbox{promptbox}[1]{
  enhanced,
  boxrule=0.8pt,
  colback=black!3,
  colframe=black!60,
  arc=2mm,            
  left=6pt,right=6pt,top=6pt,bottom=6pt,
  title={#1},
  coltitle=black,
  fonttitle=\bfseries,
  attach title to upper,
}

\section{Reproducibility Statement}
\revtext{To support reproduction, we will release \texttt{Bagpiper}-Base and the post-SFT checkpoint, the pre-training rich captions, and our training and evaluation recipes and scripts.}\revtag{xDgL, Q1}

\section{Related Work (Extended)}\label{app:related_work}
\textbf{Open-Ended Task Solving.}
The paradigm of open-ended task solving has matured significantly in the text domain \citep{hurst2024gpt, yang2025qwen3, olmo2025olmo, grattafiori2024llama}, where generalization is initially achieved through massive multi-task supervision. As the milestone, text foundational models like FLAN-T5 \citep{flant5} enumerate over 1,800 distinct text tasks to facilitate zero-shot generalization on unseen user requests. However, replicating this "task scaling" strategy in the audio domain is prohibitively expensive and practically infeasible due to the massive human labor and scarcity of well-defined, diverse enough audio tasks. Consequently, existing audio language models that attempt naive multi-tasking \citep{abouelenin2025phi, wang2024speechx, yang2023uniaudio, tang2023salmonn} still struggle with flexible, compositional instructions that fall outside their training distribution, as in \S\ref{exp_sft_gen}. In contrast, \texttt{Bagpiper} bypasses the need for massive task enumeration. By establishing rich captions as a universal semantic proxy, we effectively offload the burden of open-ended reasoning to the text capabilities of the LLM, enabling the solution of audio tasks without explicit audio-domain task supervision.

\textbf{Rich Captions as a Modal Interface.}
The utilization of automated, detailed captions has gained traction in recent research. Emerging works have demonstrated that descriptive captions can significantly enhance specific downstream capabilities, such as in music understanding \citep{ghosh2025music}, text-to-speech \citep{diwan2025scaling}, and text-to-audio/music generation \citep{zhu2025cosyaudio, chen2025fusionaudio, sun2024both}, with the research scope narrowed to task-specific objectives, relatively small scale, and unidirectional. \rev{5Apc, R3}{Our rich-caption interface also follows a line of contrastive and generative audio-captioning models \citep{laionclap2023, ma2025omni, af3}.} To our knowledge, \texttt{Bagpiper} is among the first frameworks to establish a universal mapping between natural language and all audio categories, and subsequently becomes a firm foundation for downstream open-ended processing.

\textbf{Universal Audio Processing.}
While the pursuit of universal audio understanding has yielded significant progress \citep{abouelenin2025phi, tang2023salmonn, af3}, the landscape of audio generation remains largely fragmented. Current state-of-the-art generation models are typically confined to specialized verticals, optimizing exclusively for speech \citep{peng2025vibevoice, ju2024naturalspeech, eskimez2024e2, chen2025f5}, environmental sound \citep{lee2025etta, hung2024tangoflux, huang2023make}, or music \citep{delay}. Even hybrid domains like singing voice synthesis, which inherently blend vocals and accompaniment \citep{yuan2025yue, lam2025analyzable}, remain restricted to their specific modality. Although pioneering works like \citet{yang2023uniaudio} have attempted to cover the full auditory spectrum, they rely on rigid and pre-defined control tokens\rev{5Apc, R2}{, extending a broader line of joint audio--text language models \citep{audiolm, audiopalm, anygpt, nextgpt, m2omni}}. \rev{JEWz, R1}{Among unified models, UALM \citep{tian2025ualm} is closest in objective, likewise unifying audio understanding and generation in one model, while agentic systems such as AudioGPT \citep{audiogpt} reach comparable breadth by orchestrating external audio tools around an LLM.} \rev{5Apc, R2}{Concurrent efforts \citep{uniaudio2, audioomni} pursue similar unification.} In contrast, \texttt{Bagpiper} is capable of synthesizing integrated acoustic scenes containing arbitrary, overlapping combinations of speech, music, and sound events within a single clip. \rev{xDgL, R3; JEWz, R2}{On the evaluation side, MMAU-Pro \citep{mmaupro} broadens audio-understanding assessment to spatial, multi-audio, and long-form inputs,} \rev{5Apc, Q7}{and ART \citep{art2026} is a complementary benchmark targeting audio-reasoning capabilities alongside AIR-Bench, AudioBench, and MMAU.}

\section{Rich Caption Examples}
\label{app:rich_caption_example}

\begin{promptbox}{\textbf{Rich Caption Example for Music}}

\rule{\linewidth}{0.55pt}
\scriptsize
The audio clip begins abruptly, launching into a high-energy, distorted electric guitar riff characteristic of heavy metal or hard rock, immediately accompanied by a powerful, driving drum beat. The guitar, panned hard left and treated with heavy distortion and reverb, delivers a rapid, repetitive motif that anchors the track's intensity. The drums, centered in the mix, feature a punchy kick and snare with a prominent crash cymbal, while the bass guitar sits low and centered, reinforcing the riff's rhythmic foundation. The overall sound is dense, loud, and saturated with analog-style distortion, suggesting a vintage or intentionally lo-fi production style.

At the three-second mark, a male vocalist-delivering a strained, aggressive, and raspy performance--shouts, "Sold my car!" in a non-rhotic accent reminiscent of British or Australian English. His voice, heavily processed with reverb and distortion, is panned slightly to the right and stands out against the thick instrumental backdrop. The shout is forceful and emotionally charged, with the final syllable ("car") trailing off and blending into the reverberant mix. No other vocals, harmonies, or spoken words are present.

Immediately following the vocal line, the instruments maintain their relentless energy, with the guitar and drums continuing their unyielding rhythm. The audio ends abruptly, cutting off mid-beat without any fade-out or resolution, leaving the listener with a sense of unresolved intensity.

This audio clip features a raw, high-energy rock/metal excerpt with a distorted guitar riff, driving drums, and a shouted male vocal line. The production is intentionally lo-fi, evoking the sound of 1980s underground or DIY recordings, and the lyrics "Sold my car!" are delivered with aggressive emotion. The clip is likely an excerpt from a demo or independent release, designed to convey urgency, rebellion, and personal upheaval, and is devoid of additional spoken content or environmental noise."
\end{promptbox}

\begin{promptbox}{\textbf{Rich Caption Example for Sound Effects}}

\rule{\linewidth}{0.6pt}
\scriptsize
The audio begins with the immersive, high-fidelity sound of a steady stream of water cascading over rocks in a natural outdoor setting. The water's movement is characterized by a continuous, low-frequency gurgle interwoven with midrange bubbling and a lively, high-frequency hiss, all centered in the stereo field and closely miked. This creates a richly detailed and enveloping sonic texture, evoking the presence of a small stream or brook flowing through a rocky bed, with no extraneous environmental sounds or human activity. The recording remains uninterrupted, maintaining its clarity and presence throughout, until it is abruptly and completely cut off at the end with no residual noise or fade-out.

This audio presents a pure, uninterrupted sample of natural water sounds, captured with professional equipment in a pristine outdoor environment. The absence of speech, music, or any cultural or geographic markers, along with the sudden termination, suggests a utilitarian purpose--likely as a sound effect for media production, relaxation, or ambient use. The high-quality, immersive recording is designed to evoke tranquility and natural serenity, serving as a versatile and context-free sonic backdrop.
\end{promptbox}

\begin{promptbox}{\textbf{Rich Caption Example for Speech}}

\rule{\linewidth}{0.6pt}
\scriptsize
The audio begins in a large, open urban environment, likely a city street or plaza in the Netherlands, characterized by a persistent low-frequency rumble from traffic and a faint, high-frequency hiss typical of a mobile recording device. Immediately, a sharp, metallic two-tone car horn sounds nearby, lasting about a second. The horn's clarity and proximity suggest the microphone is close to the source, and the lack of reverberation confirms an outdoor, non-enclosed setting.

As the horn fades, a female voice emerges in the background, speaking Dutch with a clear, neutral tone. She says, "Nee, maar ik ben..." ("No, but I am..."), her words partially obscured by the ongoing traffic noise. The male speaker, positioned closer to the microphone, responds succinctly with "Ja" ("Yes"), his voice calm and matter-of-fact. Both voices are slightly muffled and distant, indicating their location within the larger ambient soundscape.

The female speaker continues with a casual remark: "Laat maar even zien hoe het is dan." ("Just let me see how it is then."), delivered in a friendly, conversational manner. The male speaker replies in a relaxed, explanatory tone, "Ik vind het leuk om geluiden op te nemen." ("I like to record sounds."), directly addressing the recording's purpose. Their interaction is informal and natural, reflecting an everyday exchange in a Dutch urban context.

Throughout the clip, the background remains dominated by traffic, with no evidence of additional people, music, or environmental sounds. The conversation is unhurried and the speakers' voices are neither urgent nor emotional, indicating a casual, unscripted moment. The audio concludes abruptly, with the recording ending mid-phrase, suggesting intentional capture or editing.

In summary, this recording presents a slice of urban Dutch life: a brief, friendly exchange between two individuals, set against the ambient noise of city traffic. The male speaker explains his interest in capturing sounds, while the female speaker engages in casual conversation. The audio's technical quality, conversational flow, and environmental context reflect a spontaneous, everyday interaction in a public urban space, with no music or extraneous events present.
\end{promptbox}

\newpage
\section{Supervised Fine-Tuning (SFT) Examples}
\label{app:sft_example}

\begin{figure}[h]
    \centering
    \includegraphics[width=0.9\linewidth]{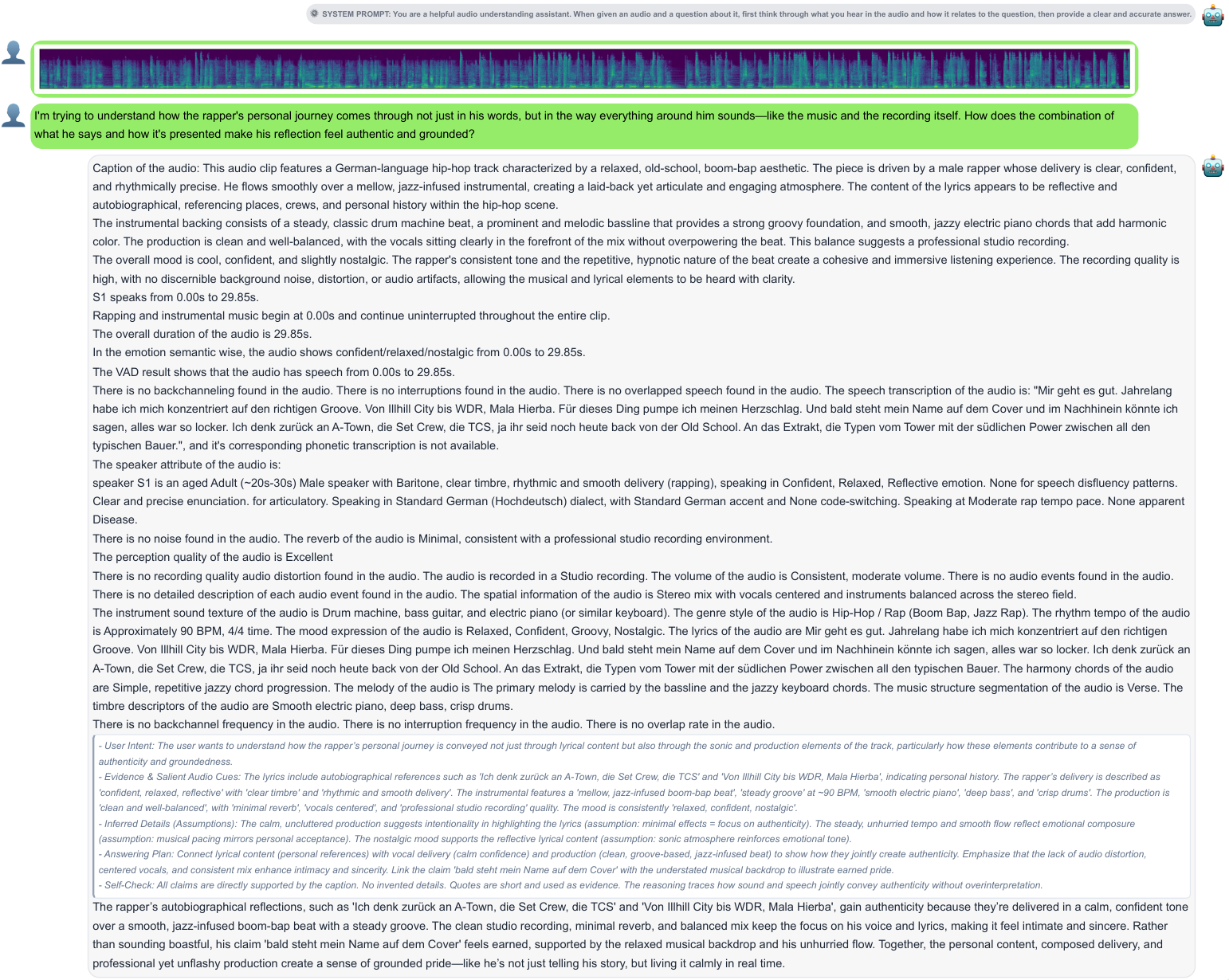}
    \caption{\textbf{SFT example (audio-to-text).} Representative audio understanding samples synthesized by our Caption-then-Process pipeline (\textit{[Audio, User Request, Rich Caption, CoT, Answer]}).}
\end{figure}
\begin{figure}[h]
    \centering
    \includegraphics[width=0.9\linewidth]{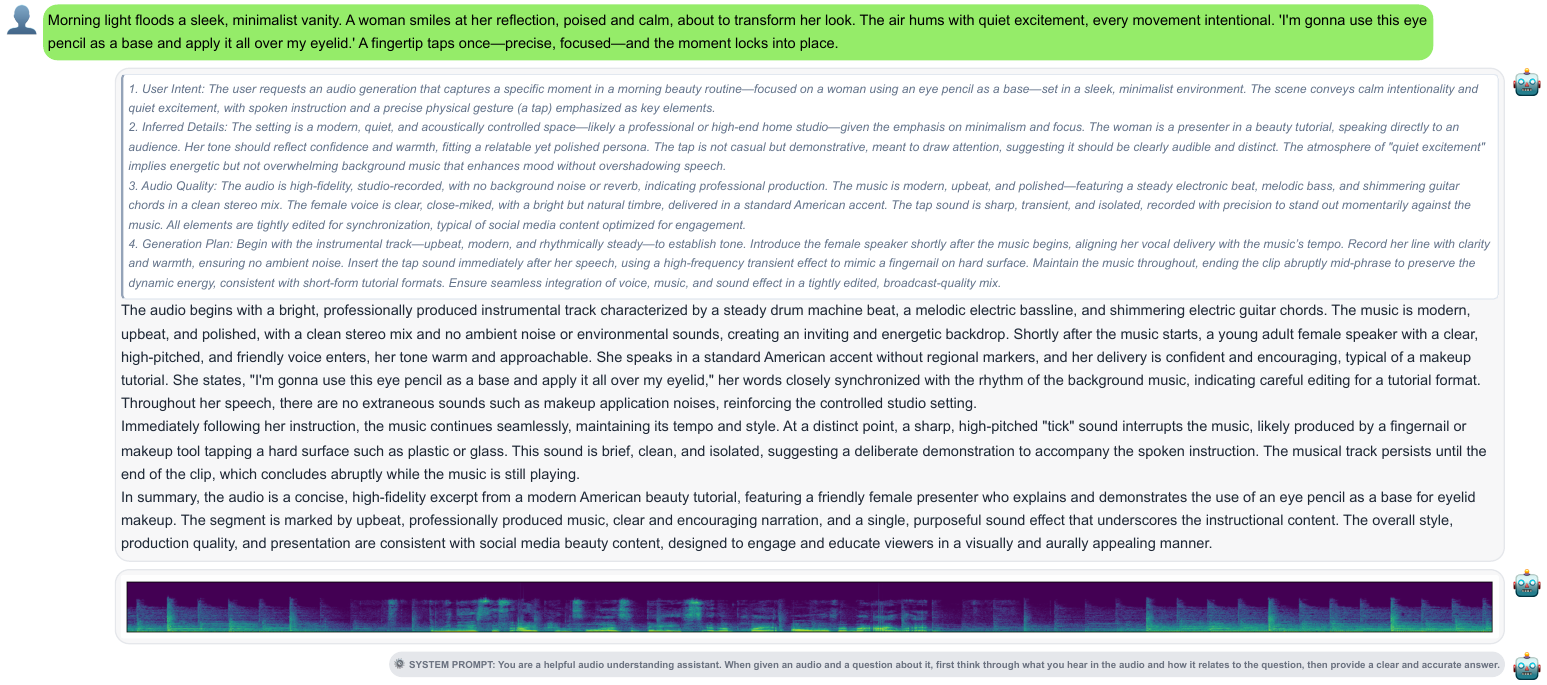}
    \caption{\textbf{SFT example (text-to-audio).} Representative audio generation samples synthesized by our simulation pipeline (\textit{[User Request, CoT, Rich Caption, Audio]}).}
\end{figure}

\newpage
\section{Implementation Details}

\subsection{Data Collection} \label{app:data_list}

\textbf{Warm-up and Pre-training Stage}: we jointly train on a mixture of audio--text paired datasets and text-only datasets.
The audio-text paired datasets include:
\textit{
OWSM v4 captions\footnote{\url{https://huggingface.co/datasets/espnet/yodas_owsmv4}},
LAION-Audio-300M\footnote{\url{https://huggingface.co/datasets/laion/LAION-Audio-300M}},
Clotho-AQA\footnote{\url{https://huggingface.co/datasets/lmms-lab/ClothoAQA}},
clotho,
Emilia\footnote{\url{https://huggingface.co/datasets/amphion/Emilia-Dataset}},
LAION captioned AI music snippets\footnote{\url{https://huggingface.co/datasets/laion/captioned-ai-music-snippets}},
LAION in-the-wild sound events\footnote{\url{https://huggingface.co/datasets/laion/in-the-wild-sound-events}},
AudioCaps\footnote{\url{https://huggingface.co/datasets/d0rj/audiocaps}},
AudioSet\footnote{\url{https://huggingface.co/datasets/agkphysics/AudioSet}},
WavCaps\footnote{\url{https://huggingface.co/datasets/cvssp/WavCaps}},
FMA\footnote{\url{https://huggingface.co/datasets/benjamin-paine/free-music-archive-full}},
YouTube-8M\footnote{\url{https://huggingface.co/datasets/leee99/yt8m-h264}},
LAION-DISCO-12M\footnote{\url{https://huggingface.co/datasets/laion/LAION-DISCO-12M}}
YODAS\footnote{\url{https://huggingface.co/datasets/espnet/yodas}}
}.
For text-only data, we use a broad \textit{Dolma3 ingredient1}\footnote{\url{https://huggingface.co/datasets/allenai/dolma3\_pool}} mixture (domain-balanced CommonCrawl high-quality subsets, plus code/math/reasoning/STEM-centric subsets) together with dialogue-style instruction corpora (\textit{llama-nemotron}\footnote{\url{https://huggingface.co/datasets/nvidia/Llama-Nemotron-Post-Training-Dataset}}, \textit{Olmo3}\footnote{\url{https://huggingface.co/collections/allenai/olmo-3}}).

\textbf{Supervised Fine-tuning Stage}: we supervise multi-turn user requests using dialogue-style instruction data that contains both audio-to-text understanding and text-to-audio generation samples (examples are shown in App.~\ref{app:sft_example}).

\begin{figure}[h]
    \centering
    \includegraphics[width=0.7\linewidth]{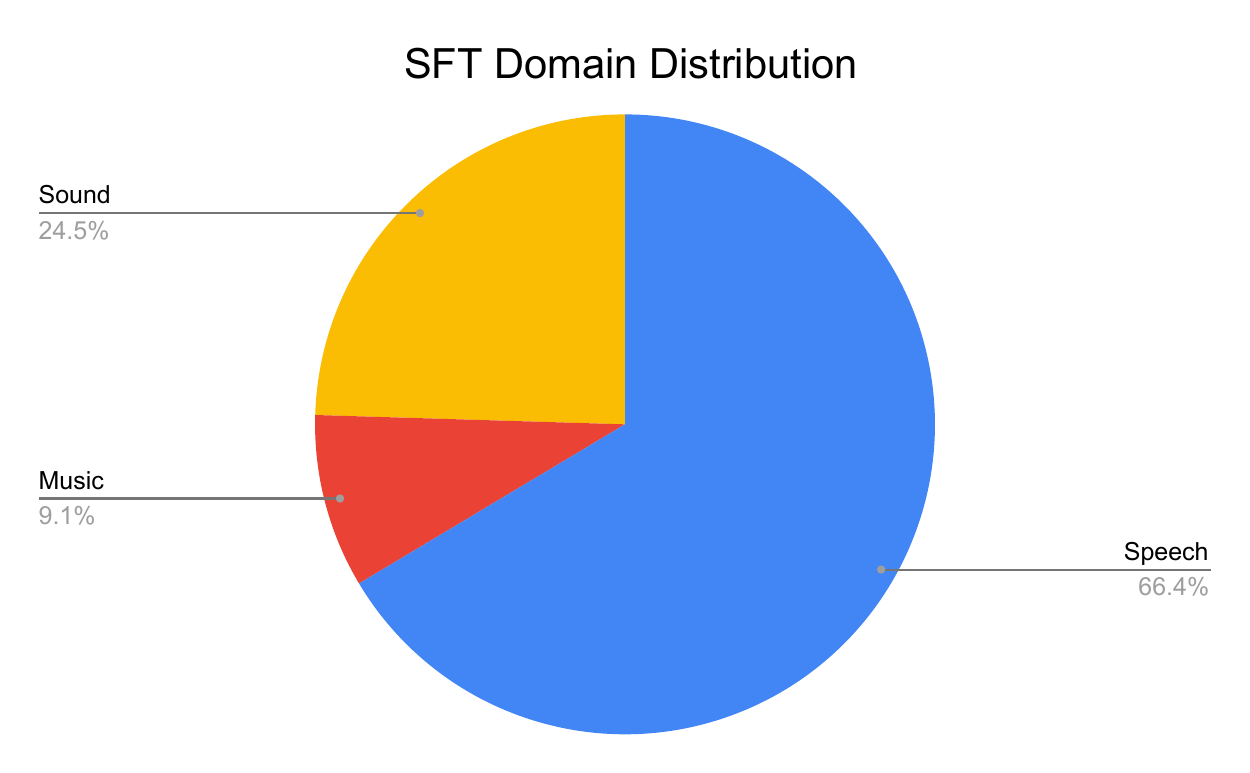}
    \caption{SFT domain distribution for speech, sound and music.}
    \label{fig:SFT_domain_distribution}
\end{figure}

The audio-to-text understanding datasets include:
\textit{
OWSM v4 (200k subset)},
\textit{Librispeech (train-clean-100), Common Voice Corpus 15, IEMOCAP, MELD, SLUE (SQA, NEL, DAC), VocCeleb1, Fake-or-Real}, \textit{Free Music Archive (FMA)
}, \textit{MUSIC-AVQA, MusicCaps, NSynth, AudioCaps, AudioGrounding, AVQA, Clotho-AQA, CochlScene}, \textit{2025 DCASE AudioQA
}, and \textit{SoundDescs
}, \textit{TUT-acoustic-scenes-2017, and VocalSound}.
We construct the SFT training set by excluding all test splits from our evaluation benchmarks. \revtext{Test-split exclusion does not prevent the same underlying audio from entering through another parent corpus. Using audfprint\footnote{\url{https://github.com/dpwe/audfprint}}, we find that 12.8\% of AudioBench and 3.2\% of AIR-Bench evaluation clips overlap with the SFT pool. The current scores, particularly AudioBench, should therefore be interpreted with this caveat.}\revtag{5Apc, R6; AC, \#2} The overall training split is formed by randomly sampling all valid samples. The resulting data composition is summarized in Table~\ref{tab:sft_dataset_counts}, showing broad coverage across speech, music, and general sound domains. The distribution of three domains is also presented in Figure~\ref{fig:SFT_domain_distribution}.

To standardize input length, when a source corpus contains clips longer than 30 seconds, we extract a 30-second segment by truncating from a uniformly random start time within the clip. In contrast, for multimodal datasets such as AVQA, where audio is tightly aligned with video content, random truncation may break semantic coherence. We therefore discard videos longer than 30 seconds and retain only those shorter than 30 seconds long. After this filtering, 50 AVQA videos remain; applying our per-corpus random sampling then yields 7 AVQA samples in the final SFT dataset. Besides, in our SLUE dataset, we only include the training set in SQA, NEL and DAC splits, since all audios in the TED split are all longer than 30 seconds. 

For text-to-audio generation data, we use a subset of all the pre-training data. Specifically, we use Gumbel Top-k sampling to sample 1M samples from the pre-training distribution, with the temperature of 0.03.

\begin{table}[t]
\centering
\caption{Dataset distribution for the SFT stage captioned by Gemini\_2.5\_Pro and Gemini\_3\_Flash, grouped by domain. ``--'' indicates the dataset is not present for that model. 
}
\small
\begin{tabular}{llrrr}
\hline
\textbf{Domain} & \textbf{Dataset} & \textbf{Gemini\_2.5\_Pro} & \textbf{Gemini\_3\_Flash} & \textbf{All sample number} \\
\hline

\multirow{7}{*}{Speech}
& Common Voice Corpus 15   & 9996  & 23775 & 33771 \\
& IEMOCAP       & 7869  & --    & 7869  \\
& Librispeech (train-clean-100)   & 10000 & 4368  & 14368 \\
& MELD          & 9984  & --    & 9984  \\
& SLUE          & --    & 12779 & 12779 \\
& VoxCeleb1     & 10000 & 81144 & 91144 \\
& Fake-or-Real  & 9994  & --   & 9994  \\
\cline{2-5}
& \textbf{Speech subtotal} & \textbf{57843} & \textbf{122066} & \textbf{179909} \\
\hline

\multirow{4}{*}{Music}
& FMA           & 1813 & --    & 1813  \\
& MUSIC-AVQA     & 50   & --    & 50    \\
& MusicCaps     & 2644 & --    & 2644  \\
& NSynth        & 9993 & 10045 & 20038 \\
\cline{2-5}
& \textbf{Music subtotal} & \textbf{14500} & \textbf{10045} & \textbf{24545} \\
\hline

\multirow{9}{*}{Sounds}
& AudioCaps           & 10000 & 8793 & 18793 \\
& AudioGrounding      & 3528  & --   & 3528  \\
& AVQA                & --    & 7    & 7     \\
& Clotho-AQA           & 1169  & --   & 1169  \\
& CochlScene          & 9993  & --   & 9993  \\
& 2025\_DCASE\_AudioQA  & 9314  & --   & 9314  \\
& SoundDescs          & 5629  & --   & 5629  \\
& TUT-acoustic-scenes-2017  & 3508  & --   & 3508  \\
& VocalSound          & 9996  & 4482 & 14478 \\
\cline{2-5}
& \textbf{Sounds subtotal} & \textbf{53137} & \textbf{13282} & \textbf{66419} \\
\hline

\multicolumn{2}{l}{\textbf{Sum}} & \textbf{125480} & \textbf{145393} & \textbf{270873} \\
\hline
\end{tabular}
\label{tab:sft_dataset_counts}
\end{table}

\newpage
\subsection{Data Curation} \label{app:curation}
We provide additional details for data curation, as mentioned in \S\ref{method_pt}.

\textbf{Audio.}
For perceptual quality, we use UTMOS for speech and the AudioBox aesthetic score for non-speech audio.

\textbf{text.} We have rule-based filtering that excludes: 
(1) labeled incomplete in rich captioner captioning; 
(2) contain non-Latin characters; 
(3) character repetition for 5 times or above;
(4) word repetition for 3 times or above;
(5) 5-gram repetition for 10\% and above;
(6) unique word ratio for 30\% or below;
(7) average word length lower than 2 or larger than 15;
(8) number of tokens larger than 800 or smaller than 200.

We additionally prompt the \texttt{Qwen3-32B} to:
(1) find the audio type, a.k.a., speech, music, sound effects;
(2) find if the text is purely in English, and exclude it if not;
(3) find the text is centralized to audio content, and exclude it if not;
(4) score the audio intelligibility, with a scale of 1--5;
(5) score the audio complexity, with a scale of 1--5;
(6) score the audio diversity, with a scale of 1--5;

\textbf{Audio-Text Alignment.} 
We use CLAP model to test the audio-text alignment. As the CLAP model would only accept text up to 77 tokens, we use \texttt{Qwen3-32B} to summarize it into short captions.

\textbf{Gumbel Top-K sampling.}
For audio understanding, we use a temperature of 0.3/0.3/0.3 and discard 20\%/20\%/42\% examples for sound effects/music/speech.
For audio generation, we use a temperature of 0.1/0.1/0.1 and discard 20\%/20\%/28\% examples for sound effects/music/speech.

\subsection{Model Details}

The model details discussed in Sec.\ref{sec:arch}, are shown in Table~\ref{app:model_details}

\begin{table}[h]
\centering
\caption{Bagpiper Model configuration.}
\label{app:model_details}
\begin{tabular}{lc}
\toprule
{Component} & {Model Configuration}  \\
\midrule
Text backbone & Qwen/Qwen3-8B-Base  \\
Discrete audio tokenizer & X-Codec (hubert-general)  \\
Continuous audio encoder & Qwen/Qwen3-Omni-30B-A3B-Instruct-Encoder  \\
\bottomrule
\end{tabular}
\end{table}

\subsection{Training Details (3-stages)}

We train the model in three stages: (1) warmup, (2) large-scale pretraining, and (3) supervised fine-tuning (SFT).
Across all stages, we use packed batching (\texttt{batchfy\_method=pack}), mixed-modality data, and DeepSpeed for distributed optimization.
All stages use \texttt{dtype=bfloat16} with activation checkpointing enabled. The pre-training is complete on 80 NVIVIDA GH200 GPUs.

\begin{table}[h]
\caption{Training configuration across the three stages. The pipeline progresses from connector alignment to full backbone pre-training, concluding with supervised fine-tuning (SFT).}
\centering
\label{tab:training_details}
\begin{tabular}{lccc}
\toprule
 & Warmup & Pre-train & SFT \\
\midrule
Frozen Modules & Audio Encoder, Transformer Body & Audio Encoder & Audio Encoder \\
Batch Size (Token) & 4.8M & 1.2M & 1.2M \\
Optimizer & AdamW & AdamW & AdamW \\
LR Schedule & Constant & Cosine Decay & Constant \\
Warmup Steps & 200 & 5k & - \\ 
Steps & 2k & 480k & 12.5k \\
Peak LR & $5 \times 10^{-4}$ & $1 \times 10^{-4}$ & $1 \times 10^{-5}$ \\
\bottomrule
\end{tabular}
\end{table}

\newpage
\subsection{Inference Configuration}

We employ modality-specific decoding with \texttt{bfloat16} precision and a greedy search ($N=1$), other details shown in Table~\ref{app:inference_details}.
\rev{JEWz, Q4}{\texttt{Bagpiper} has around 8B parameters and is served in BF16, so the model weights take roughly 16GB of GPU memory. In practice, a GPU with around 24GB memory or above can run inference smoothly, leaving space for activations and decoding buffers.}

\begin{table}[h]
\centering
\caption{Inference configuration with modality-specific decoding hyper-parameters.}
\begin{tabular}{l cccc}
\toprule
{Item} & {Temperature} & Top-$k$ & Classifier-Free Guidance~(CFG) & max-decode-steps \\
\midrule
Text decoding & 0.6 & 20 & - & 2048 \\
Audio decoding & 0.8 & 20 & 3 & 2048 \\
\bottomrule
\end{tabular}
\label{app:inference_details}
\end{table}

\newpage
\subsection{Model Roles}\label{app:model_roles}

\revtext{Table~\ref{tab:model_roles} lists every model used in \texttt{Bagpiper}, grouped by the role it plays. We separate three kinds of role: the \emph{components} that constitute the model itself, the \emph{data-pipeline} models used to caption, classify, simulate, and filter training data, and the \emph{evaluation} models used to score results. The pipeline is organised so that no evaluation judge contributes to the training data it later scores. In the understanding track, rich captions come from \texttt{Gemini}, the SFT data is simulated and filtered by \texttt{Qwen3-235B-A22B} and \texttt{GPT-OSS-120B}, and evaluation uses verifiable metrics together with \texttt{GPT-4o}. In the generation track, rich captions come from the \texttt{Qwen3-Omni-Captioner}, the simulators are the same, and the A/B test is judged by \texttt{Gemini-3-Pro} and \texttt{GPT-4o}. In each track the judges are therefore disjoint from the models that produced the training data. The pre-training probe is likewise disjoint: pre-training captions come from the \texttt{Qwen3-Omni-Captioner}, while the probing LLM is \texttt{Gemini-3-Flash}. The only model appearing on both sides of the table is \texttt{Gemini-3-Pro}, which produced the attribute taxonomy used to diversify \emph{understanding} prompts and judges \emph{generation} outputs; these are different tracks, so it never scores data it helped create.}\revtag{5Apc, R4}

\begin{table}[h]
\centering
\caption{Models used in \texttt{Bagpiper} and their roles.\revtag{5Apc, R4}}
\label{tab:model_roles}
\small
\revtext{
\begin{tabular}{llc}
\toprule
\textbf{Role} & \textbf{Model} & \textbf{Where} \\
\midrule
\multicolumn{3}{l}{\textit{Model components}} \\
LLM backbone                        & \texttt{Qwen3-8B-Base}                          & \S\ref{sec:arch} \\
Continuous audio encoder            & \texttt{Qwen3-Omni-30B-A3B-Instruct} encoder    & \S\ref{sec:arch} \\
Discrete audio codec                & \texttt{X-Codec}                                & \S\ref{sec:arch} \\
\midrule
\multicolumn{3}{l}{\textit{Pre-training data pipeline}} \\
Rich captioner                      & \texttt{Qwen3-Omni-30B-A3B-Captioner}           & \S\ref{method_pt} \\
Audio taxonomy classifier           & \texttt{Qwen3-32B}                              & \S\ref{method_pt} \\
Text-quality rubric judge           & \texttt{Qwen3-32B}                              & \S\ref{app:curation} \\
Caption summariser (for CLAP)       & \texttt{Qwen3-32B}                              & \S\ref{app:curation} \\
Audio-quality scorers               & UTMOS; audiobox-aesthetics                      & \S\ref{method_pt} \\
Audio--text alignment scorer        & CLAP                                            & \S\ref{method_pt} \\
\midrule
\multicolumn{3}{l}{\textit{SFT data pipeline}} \\
Rich captioner (understanding)      & \texttt{Gemini-3-Flash}; \texttt{Gemini-2.5-Pro} & \S\ref{method_sft} \\
Rich captioner (generation)         & \texttt{Qwen3-Omni-30B-A3B-Captioner}           & \S\ref{method_sft} \\
Attribute taxonomy extraction       & \texttt{Gemini-3-Pro}                           & \S\ref{sft_pipeline} \\
Request / CoT simulator             & \texttt{Qwen3-235B-A22B-Instruct-2507-FP8}; \texttt{GPT-OSS-120B} & \S\ref{sft_pipeline} \\
Training-data quality filter        & \texttt{Qwen3-235B-A22B-Instruct-2507-FP8}      & \S\ref{sft_pipeline} \\
\midrule
\multicolumn{3}{l}{\textit{Evaluation}} \\
Pre-training probing LLM            & \texttt{Gemini-3-Flash}                         & \S\ref{exp_pt} \\
Understanding judge                 & \texttt{GPT-4o}                                 & \S\ref{exp_sft_und} \\
Generation A/B judge                & \texttt{Gemini-3-Pro}; \texttt{GPT-4o}          & \S\ref{exp_sft_gen} \\
\bottomrule
\end{tabular}}
\end{table}

\newpage
\subsection{SFT Filter Validation}\label{app:filter_validation}

\revtext{The SFT quality filter of \S\ref{sft_pipeline} retains samples whose average score over the five rubric dimensions exceeds 3. To check that this threshold is not cutting into the central mass of the score distribution, Table~\ref{tab:filter_validation} reports the quartiles of the per-example average score together with the resulting retention rate. Under the \texttt{Qwen3-235B-A22B} filter the first quartile coincides with the threshold ($p_{25}=3.00$) and about $75\%$ of the pool is retained, so the cutoff removes the lowest quartile rather than samples near the midpoint. As a cross-family check we scored the same pool with \texttt{GPT-OSS-120B}; its scores saturate near the top of the scale and it retains the entire pool, giving little discrimination on this fine-grained rubric. This is why \texttt{Qwen3-235B-A22B} is used as the filter in the main pipeline.}\revtag{5Apc, Q4}

\begin{table}[h]
\centering
\caption{Validation of the SFT filter cutoff. Quartiles are over per-example average scores across the five rubric dimensions; retention is the fraction of the pool kept at the ``average $>3$'' threshold.\revtag{5Apc, Q4}}
\label{tab:filter_validation}
\revtext{
\begin{tabular}{lccc}
\toprule
Judge & Avg. score ($p_{25}$ / $p_{50}$ / $p_{75}$) & Retained & Discrimination \\
\midrule
\texttt{Qwen3-235B-A22B} & 3.00 / 3.19 / 3.38 & 75.0\% & informative \\
\texttt{GPT-OSS-120B}    & 4.75 / 4.88 / 4.94 & 100\%  & saturated \\
\bottomrule
\end{tabular}}
\end{table}

\subsection{Caption-Error Propagation}\label{app:error_propagation}

\revtext{We examine whether errors in the intermediate rich captions produced by the fine-tuned \texttt{Bagpiper} of \S\ref{exp_sft_und} and \S\ref{exp_sft_gen} propagate to its downstream predictions. For understanding, we evaluate the 333-example speech subset of MMAU-Mini. \texttt{Bagpiper} first generates rich captions from audio; \texttt{Qwen3-235B-A22B-FP8} then revises these captions using pseudo-ground-truth transcriptions produced by \texttt{Qwen3-ASR-1.7B}, after which \texttt{Bagpiper} performs the final reasoning from the revised captions. Correcting the transcriptions raises accuracy only from $72.4\%$ to $73.0\%$, suggesting that ASR errors have limited impact on this subset. For generation, we evaluate 2,620 LibriSpeech test-clean examples. We use \texttt{Qwen3-235B-A22B-FP8} to extract the transcription from each intermediate rich caption and compare it with the text in the original user request. The resulting WER is $0.2\%$, showing that the rich-caption step preserves the requested text with high fidelity in this TTS-style setting; most remaining errors therefore arise in the subsequent caption-to-audio stage.}\revtag{xDgL, R2}

\begin{table}[h]
\centering
\caption{Propagation of errors through the intermediate rich caption.\revtag{xDgL, R2}}
\label{tab:error_propagation}
\revtext{
\begin{tabular}{lll}
\toprule
Analysis & Data & Result \\
\midrule
ASR error $\rightarrow$ understanding & MMAU-Mini speech (333) & $72.4\% \rightarrow 73.0\%$ accuracy \\
Caption error $\rightarrow$ generation & LibriSpeech test-clean (2,620) & $0.2\%$ WER \\
\bottomrule
\end{tabular}}
\end{table}

\subsection{Audio-Understanding Ablations}\label{app:understanding_ablations}

\revtext{Table~\ref{tab:understanding_ablations} consolidates the controlled ablations reported during rebuttal. Applying the same SFT recipe directly to \texttt{Qwen3-8B-Base} without audio pre-training degrades all four understanding metrics and, in our qualitative evaluation, does not produce intelligible audio. The CoT comparison isolates the contribution of the caption-plus-reasoning training format under this setup. The backbone-scale comparison holds the training recipe and 50k-step budget fixed; neither row is the final checkpoint. For the pipeline baseline, a frozen \texttt{Qwen3-Omni-30B-A3B-Captioner} receives audio without a text prompt, then an instruction-tuned \texttt{Qwen3-8B} receives only the resulting caption, question, and answer options. The text LLM has no access to the audio. Despite using this larger perception frontend, the pipeline trails the end-to-end 8B \texttt{Bagpiper} by 11.3, 11.9, and 0.7 points on MMAU-Mini, MMAU, and MMAR, respectively.}\revtag{xDgL, Q2/Q4/Q5; xDgL, R4; JEWz, Q1}

\begin{table}[h]
\centering
\caption{Audio-understanding ablations. Scale variants are evaluated after 50k training steps. ``--'' denotes an unreported metric.\revtag{xDgL, Q2/Q4/Q5; xDgL, R4; JEWz, Q1}}
\label{tab:understanding_ablations}
\scriptsize
\begingroup
\setlength{\tabcolsep}{2.5pt}
\revtext{
\begin{tabular}{llcccc}
\toprule
Ablation & Variant & WER ($\downarrow$) & MMAU-Mini ($\uparrow$) & MMAU ($\uparrow$) & MMAR ($\uparrow$) \\
\midrule
Pre-training & with pre-training & 2.5 & 74.5 & 73.1 & 57.0 \\
             & SFT-only & 4.1 & 70.2 & 66.7 & 49.3 \\
\midrule
CoT format   & with CoT & -- & 74.5 & 73.1 & 57.0 \\
             & without CoT & -- & 71.0 & 67.0 & 52.2 \\
\midrule
Backbone scale & \texttt{Bagpiper} (Qwen3-8B) & -- & 68.0 & -- & -- \\
@50k steps     & \texttt{Bagpiper}-30B-A3B & -- & 74.3 & -- & -- \\
\midrule
Inference system & \texttt{Bagpiper} (end-to-end) & -- & 74.5 & 73.1 & 57.0 \\
                 & Captioner + Qwen3-8B & -- & 63.2 & 61.2 & 56.3 \\
\bottomrule
\end{tabular}}
\endgroup
\end{table}

\subsection{Attribute Taxonomy}\label{app:taxonomy}
\revtext{For completeness, we list the full set of 108 understanding attributes used to diversify SFT prompts (\S\ref{method_sft}), grouped into seven super-categories. We report only the attribute names; no per-attribute definitions or examples are used in the pipeline.}\revtag{JEWz, Q2}

{\footnotesize
\begin{multicols}{2}
\noindent\textbf{Acoustic \& Sound Perception (24).} Sound Source Identification; Sound Event Recognition; Environmental Sound Classification; Speech Content Transcription; Music Instrument Identification; Voice Identification; Animal Sound Recognition; Mechanical/Machine Sound Recognition; Natural Sound Recognition; Human Non-Speech Sounds; Musical Genre Classification; Speech vs Non-Speech Detection; Foreground/Background Sound Separation; Sound Quality Assessment; Audio Authenticity Detection; Multi-source Sound Recognition; Sound Intensity/Volume Analysis; Sound Frequency Characteristics; Audio Channel Analysis; Audio Temporal Quality; Ambient Sound Interpretation; Sound Effect Recognition; Musical Texture Analysis; Acoustic Scene Classification.

\medskip
\noindent\textbf{Speaker \& Speech Attributes (18).} Speaker Count Estimation; Speaker Gender Identification; Speaker Age Estimation; Speaker Emotion Recognition; Speaker Accent/Dialect Detection; Speaker Native Language Inference; Speaker Role/Occupation Inference; Speaker Identity Verification; Speaker Relationship Inference; Speech Pace/Rate Analysis; Speech Clarity/Articulation; Speech Fluency Analysis; Speech Formality Level; Phonemic Stress Pattern Analysis; Phonological Sequence Decoding; Multi-Speaker Role Mapping; Conversation Turn-Taking Analysis; Speech Act Classification.

\medskip
\noindent\textbf{Emotion \& Sentiment (12).} Emotion State Summarization; Emotional Tone Interpretation; Emotion Flip/Change Detection; Sarcasm/Irony Detection; Sentiment Analysis; Speaker Intention Inference; Dissonant Emotion Interpretation; Genuine vs Fake Emotion Detection; Urgency/Importance Detection; Mood Induction Analysis; Threatening vs Non-threatening Classification; Excitement/Enthusiasm Level.

\medskip
\noindent\textbf{Temporal \& Rhythmic (14).} Event Counting; Event Duration Estimation; Event Ordering/Sequencing; Temporal Event Reasoning; Sound Onset/Offset Detection; Rhythm and Tempo Understanding; Time Signature Detection; Beat Counting; Temporal Alignment Analysis; Event Frequency Estimation; Duration Comparison; Sequence Prediction; Temporal Anomaly Detection; Event Timeline Construction.

\medskip
\noindent\textbf{Spatial \& Environmental (10).} Spatial Sound Localization; Indoor/Outdoor Environment Detection; Room/Space Type Inference; Distance/Proximity Estimation; Sound Movement Direction; Environmental Perception and Reasoning; Background Environment Inference; Acoustic Space Characteristics; Reverberation/Echo Analysis; Multi-location Sound Mapping.

\medskip
\noindent\textbf{Content \& Semantic (18).} Factual Information Extraction; Conversational Fact Retrieval; Key Highlight Extraction; Topic/Subject Identification; Main Idea Summarization; Detail Extraction; Lyrical Content Analysis; Spoken Word Comprehension; Dialogue Understanding; Narrative Structure Analysis; Argumentation Analysis; Question Answering from Speech; Instruction Comprehension; Named Entity Recognition; Numerical Information Extraction; Relationship Extraction; Causal Reasoning; Inference/Implication.

\medskip
\noindent\textbf{Knowledge \& Cultural (12).} Cultural Context Recognition; Music Theory Knowledge; Professional Domain Knowledge; Historical/Cultural Event Association; Language/Dialect Culture; Musical Genre Cultural Context; Social Convention Understanding; Domain-Specific Terminology; Geographic/Regional Inference; Era/Period Detection; Style/Artistic Movement Recognition; Cross-Cultural Comparison.
\end{multicols}
}

\section{Additional Generation Evaluations}\label{app:additional}

\subsection{Matched-Input Control for Open-Ended Generation}\label{app:matched_input}
\revtext{In the main-text A/B test (Fig.~\ref{fig:win_rate_results}), only the baselines receive the condensed 15-word instruction while \texttt{Bagpiper} receives the full prompt. To isolate the model contribution from this input-length difference, we additionally give \texttt{Bagpiper} the same 15-word summary (\texttt{Bagpiper}-summary) and re-run the evaluation under both judges (Table~\ref{tab:matched_input}), comparing it both against the full-prompt baselines and against the matched \textit{summary} baselines. \texttt{Bagpiper}-summary still wins in aggregate against every baseline under both judges; the margins shrink (most on \textit{sound} against \texttt{TangoFlux}-summary under \texttt{GPT-4o}, where the per-domain rate dips below $50\%$), while the aggregate win rate stays above $50\%$ against every baseline, indicating the advantage is largely not an input-length artefact. Since \texttt{Bagpiper} can still apply its caption-then-process reasoning under short prompts, this is a practical matched-input control rather than a complete ablation of the reasoning step.}\revtag{9r2o, Q3}

\begin{table}[h]
    \centering
    \small
    \caption{\revtext{Matched-input control: A/B win rates (\%) of \texttt{Bagpiper}-summary (given the same 15-word instruction as the baselines), broken down by \textbf{Sp}eech / \textbf{So}und / \textbf{Mu}sic / \textbf{All} under two judges.}}
    \label{tab:matched_input}
    \begin{tabular}{lcccccccc}
    \toprule
     & \multicolumn{4}{c}{\texttt{Gemini-3-Pro}} & \multicolumn{4}{c}{\texttt{GPT-4o}} \\
    \cmidrule(lr){2-5}\cmidrule(lr){6-9}
    \texttt{Bagpiper}-summary vs & Sp & So & Mu & All & Sp & So & Mu & All \\
    \midrule
    AudioLDM2-Large          & 99.1 & 72.2 & 81.9 & 84.4 & 94.9 & 56.5 & 65.9 & 72.5 \\
    AudioLDM2-Large-summary   & 99.7 & 74.0 & 84.0 & 85.9 & 96.4 & 63.4 & 70.4 & 76.8 \\
    TangoFlux                & 97.9 & 54.7 & 81.3 & 78.0 & 87.1 & 46.5 & 65.6 & 66.4 \\
    TangoFlux-summary         & 98.5 & 47.7 & 81.3 & 75.9 & 85.6 & 40.5 & 67.7 & 64.6 \\
    \bottomrule
    \end{tabular}
\end{table}

\subsection{TTS-Specialized Variant}\label{app:tts_specialized}
\revtext{Figure~\ref{fig:win_rate_results} measures open-ended instruction-following rather than dedicated text-to-speech quality. To compare against specialized systems on their own terrain, we fine-tune a TTS-specialized \texttt{Bagpiper} variant and run side-by-side preference tests on LibriSpeech test-clean against \texttt{Qwen3-TTS}, \texttt{CosyVoice3}, and \texttt{VibeVoice}, judged separately on \emph{overall quality} and \emph{prosody/naturalness} under both \texttt{Gemini-3-Pro} and \texttt{GPT-4o} (Table~\ref{tab:tts_specialized}). The variant is preferred over both \texttt{CosyVoice3} and \texttt{VibeVoice} under both judges and both rubrics, and is competitive with \texttt{Qwen3-TTS}: the only sub-$50\%$ cell is \texttt{GPT-4o} prosody/naturalness against \texttt{Qwen3-TTS} ($48.7\%$). In the pooled pairwise ranking the ordering is consistent across judges and rubrics: \texttt{Bagpiper} $>$ \texttt{VibeVoice} $>$ \texttt{CosyVoice3}.}\revtag{9r2o, Q1/R2}

\begin{table}[h]
    \centering
    \small
    \caption{\revtext{A/B win rates (\%) of the TTS-specialized \texttt{Bagpiper} variant on LibriSpeech test-clean, judged separately on overall quality and prosody/naturalness under two judges. Reported separately from Fig.~\ref{fig:win_rate_results}.}}
    \label{tab:tts_specialized}
    \begin{tabular}{llcc}
    \toprule
    \texttt{Bagpiper}-TTS vs & Rubric & \texttt{Gemini-3-Pro} & \texttt{GPT-4o} \\
    \midrule
    \multirow{2}{*}{\texttt{Qwen3-TTS}}  & Overall quality        & 71.3 & 57.5 \\
                                         & Prosody / naturalness  & 72.0 & 48.7 \\
    \midrule
    \multirow{2}{*}{\texttt{CosyVoice3}} & Overall quality        & 84.0 & 65.7 \\
                                         & Prosody / naturalness  & 85.7 & 57.7 \\
    \midrule
    \multirow{2}{*}{\texttt{VibeVoice}}  & Overall quality        & 61.7 & 55.3 \\
                                         & Prosody / naturalness  & 58.3 & 51.3 \\
    \bottomrule
    \end{tabular}
\end{table}

\subsection{Stage-Attributed Qualitative Study}\label{app:stage_study}
\revtext{We attribute the qualitative behaviors of \S\ref{subsec:qualitative} to the training stage that produces each.
\textbf{Base (\texttt{Qwen3-8B}).} Before audio training the base model is not an audio generator at all: it has no mechanism to emit audio tokens and produces no audio output.
\textbf{Pre-trained (\texttt{Bagpiper}-Base).} Audio pre-training endows the model with compositional audio-token generation; intelligible, well-formed audio begins to emerge at roughly $100$k pre-training steps.
\textbf{Post-SFT (\texttt{Bagpiper}).} Supervised fine-tuning makes this generative substrate controllable by open-ended natural-language requests, yielding the compositional synthesis, logic-constrained instruction following, and world-knowledge grounding observed in \S\ref{subsec:qualitative}.
The generative capability is therefore attributable to pre-training and its controllability to SFT.}\revtag{JEWz, R5}

\section{LLM Usage Disclosure}
In compliance with the COLM 2026 LLM Usage Policy, we disclose that Large Language Models (LLMs)—both locally deployed and accessed via remote APIs—were utilized for data curation, simulation, and verification tasks as detailed in $\S\ref{sec:method}$. During manuscript preparation, LLMs provided minor assistance limited to grammar refinement and rephrasing. We explicitly state that LLMs were NOT used to originate core research ideas, draft original technical prose, or generate experimental plots.

\end{document}